%% file: manuscript.tex
\documentclass[Afour,sageh,times]{sagej}

\usepackage{moreverb,url}

\usepackage[colorlinks,bookmarksopen,bookmarksnumbered,citecolor=red,urlcolor=red]{hyperref}
\usepackage{multirow}
\usepackage{makecell}
\usepackage{float}

\newcommand\BibTeX{{\rmfamily B\kern-.05em \textsc{i\kern-.025em b}\kern-.08em
T\kern-.1667em\lower.7ex\hbox{E}\kern-.125emX}}

\def\volumeyear{2016}

\usepackage{xcolor}

\input{defs}

\begin{document}

\runninghead{Ha et al.}

\title{Learning-based legged locomotion; state of the art and future perspectives}

\author{Sehoon Ha\affilnum{1}, Joonho Lee \affilnum{2}, Michiel van de Panne \affilnum{3}, Zhaoming Xie \affilnum{4}, Wenhao Yu \affilnum{5}, Majid Khadiv\affilnum{6}}

\affiliation{authors are listed alphabetically (except the last author). \\
\affilnum{1}Georgia Institute of Technology, USA\\
\affilnum{2}Neuromeka, Korea\\
\affilnum{3}University of British Columbia, Canada\\
\affilnum{4}The AI Institute, USA\\
\affilnum{5}Google DeepMind, USA\\
\affilnum{6}Technical University of Munich, Germany}

\corrauth{Majid Khadiv, Munich Institute of Robotics and Machine Intelligence (MIRMI),
Technical University of Munich (TUM), Georg-Brauchle-Ring 60-62, Munich
80992, Germany.}

\email{majid.khadiv@tum.de}

\begin{abstract}
Legged locomotion holds the premise of universal mobility, a critical capability for many real-world robotic applications.
Both model-based and learning-based approaches have advanced the field of legged locomotion in the past three decades.  
In recent years, however, a number of factors have dramatically accelerated progress in learning-based methods, including the rise of deep learning, rapid progress in simulating robotic systems, and the availability of high-performance and affordable hardware. 
This article aims to give a brief history of the field, to summarize recent efforts in learning locomotion skills for quadrupeds, and to 
provide researchers new to the area with an understanding of the key issues involved.
With the recent proliferation of humanoid robots, we further outline the rapid rise of analogous methods for bipedal locomotion. 
We conclude with a discussion of open problems as well as related societal impact.

%We conclude with a discussion of unsolved problems and open directions in the field of learning-based robot locomotion. We also briefly discuss the potential societal impacts of this technology and present some thoughts on how we can safe-guard it against potential harmful applications. 
\end{abstract}


\maketitle

\section{Introduction}
\begin{figure}
    \centering
    \includegraphics[width=.9\linewidth]{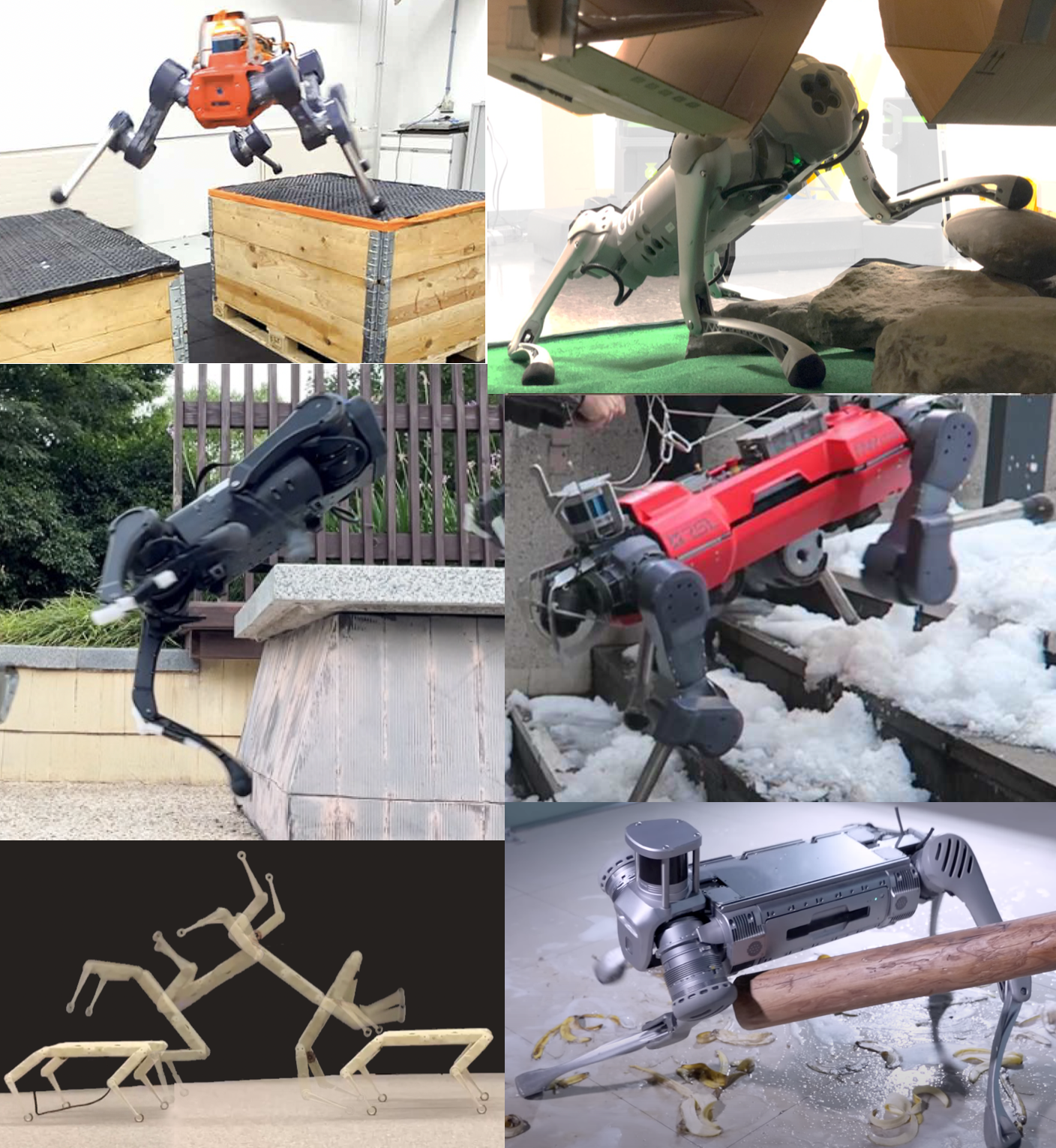}\vspace{-2mm}
    \caption{Examples of notable learned behaviors in the real world  \cite{hoeller2023anymal,xu2024dexterous,zhuang2023robot,miki2022learning,li2023learning,unitree2024b2}.}\vspace{-9mm}
    \label{fig:snapshots}
\end{figure}

Legged robots are complex systems with highly nonlinear, hybrid, and inherently unstable dynamics. Early demonstrations in the 1980s showed that simple feedback mechanisms could achieve robust dynamic locomotion for systems with one, two, and four legs~\cite{raibert1984hopping,raibert1986legged}. This set the stage for more than forty years of research on legged locomotion. In particular, we have witnessed explosive progress in quadruped locomotion over the past few years (Figure~\ref{fig:snapshots}). There are multiple reasons for this rapid progress, including reliable and affordable hardware, significant improvements in simulation environments, and scalable learning algorithms for high-dimensional continuous control problems. There has been further recent success with similar approaches for bipedal and humanoid robots.
We believe this is therefore an opportune time to broadly summarize the efforts to date as well as to reflect on future research directions. This article aims to provide current and new researchers with a big-picture view of how the field has evolved to date. This review article may also provide a useful foundation for courses covering \emph{learning-based legged locomotion}.
We note that while many learning-based approaches can be applied more generically to arbitrary robot tasks, e.g., including manipulation, we focus specifically on the challenges related to learning legged locomotion.

To place the remainder of the survey in a relevant context, 
we now provide a historical perspective of the key elements that have enabled the rapid advances seen for learning-based legged locomotion, with a focus on quadrupeds. This includes the evolution of the hardware, the physics-based simulators required for learning, and the most common methods and algorithms applied to learning-based control. We also briefly situate this survey itself with respect to other recent surveys.

\subsection{Hardware}
\begin{figure*}
    \centering
    \includegraphics[trim={.1cm 0 0 0},clip,width=.99\linewidth]{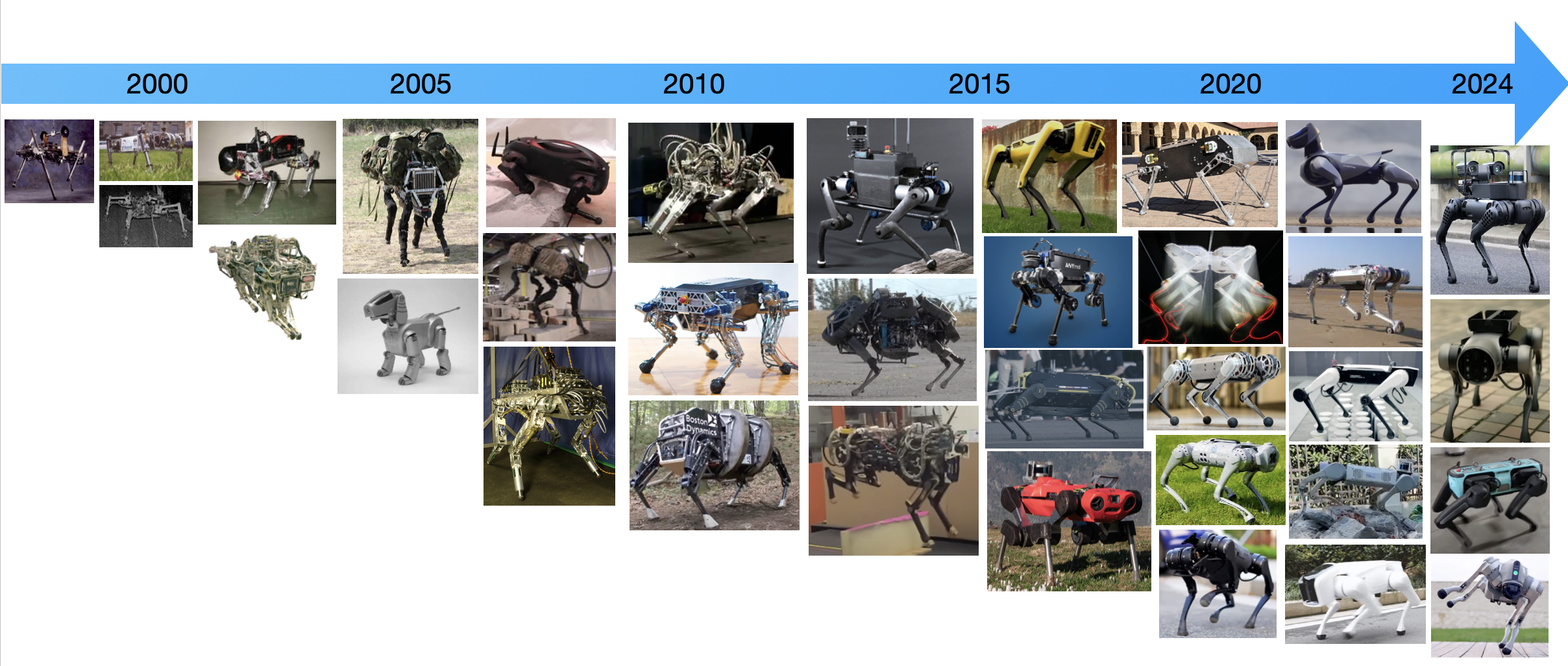}
    \caption{Evolution of quadruped hardware over time}
    \label{fig:hardware}
\end{figure*}

Hardware for quadrupedal locomotion has evolved considerably over the past few decades in terms of both capabilities and cost (Figure~\ref{fig:hardware}).
Torque-controlled quadruped robots traditionally use one of three actuation mechanisms: hydraulic, electric motors with torque sensors, and series elastic actuators. The legged robots from the Leg Lab~\cite{raibert1986legged} and later the success of the Bigdog project~\cite{Raibert2008BigDogTR} introduced hydraulics as an attractive actuation mechanism for quadrupedal locomotion. In particular, the large power-to-weight ratio enables hydraulically-actuated quadruped robots to perform highly dynamic motions and carry large payloads \cite{doi2006development,semini2010hyq,rong2012design,junyao2013research,semini2016design}. However, hydraulic systems are expensive and require specialized expertise for design, maintenance, and repair.

Electric motors are the most common choice for actuation mechanism of quadruped robots \cite{buehler1998scout,buehler1999stable}. To provide sufficient torque at the output of the joints with an electric motor, a gearing mechanism with a high ratio (such as harmonic drive) was traditionally required. However, the large static friction introduced by the gearbox renders the joints non-backdrivable which precludes output torque control by simply controlling the motor current. Classically, two modifications were introduced to enable joint torque control, namely using springs \cite{hutter2012starleth,hutter2016anymal} or joint torque/force sensors \cite{spot2015quadruped} after the gearbox. However, the former design limits the control bandwidth of the joint, and the latter is sensitive to large impact forces that could damage the gearbox or saturate (and damage) the force-torque sensor.

Using custom-made high torque-density actuators, the next generation of quadrupeds could control their interaction through direct joint current control with a low gear ratio and without a need for springs and torque sensors in the robot structure \cite{seok2014design}. This ability, sometimes called proprioceptive actuation \cite{wensing2017proprioceptive}, revolutionized the field of quadrupedal locomotion. High torque density, high-bandwidth force control, and the ability to mitigate impacts through backdrivability are some of the most interesting features of these actuators. Thanks to these features, proprioceptive actuators have become the dominant paradigm in the design of legged robots, and a wide variety of quadrupeds have been developed based on this concept \cite{katz2019mini,grimminger2020open,shin2022design}. In particular, the open-source initiatives from academia \cite{kau2019stanford,grimminger2020open} as well as inexpensive hardware from industry \cite{deeprobotics2021quadruped,unitree2021quadruped,Xiaomi2021quadruped} have dramatically accelerated experimental progress in quadrupedal locomotion. Among the many different robots, the Unitree A1 and  Go1/Go2 are among the most popular choices for showcasing learned quadrupedal behaviors in the real world, due to their price and performance.

Bipedal robots and humanoids are currently seeing resurgent interest, given that much of our world is designed around the human form factor. The hardware is following a similar evolutionary path as that seen for quadrupedal robots. Early torque-controlled humanoids often used either hydraulic actuators \cite{Smith2016,Schaal2018,Nelson2018}, electrical motors with torque sensors \cite{englsberger2014overview}, or series elastic actuators \cite{hubicki2016atrias,abate2018mechanical,ahn2019control}. A more complete summary of humanoid hardware and software can be found in \cite{goswami2018humanoid}. Recently, we are observing the same trend towards humanoids with proprioceptive actuators \cite{chignoli2021humanoid,zhu2023design}. As with quadrupeds, these systems will become more affordable and capable.   Towards the end of this survey, we revisit the key differences and opportunities of learned bipedal locomotion as compared to that for quadrupeds. 

\subsection{Simulators}\label{subsec:simulators}
The availability of highly efficient numerical algorithms for computing forward and inverse dynamics of articulated rigid body systems gave the robotics community with the necessary tools to simulate high-dimensional systems \cite{featherstone1987robot}. These algorithms can solve forward and inverse dynamics problems for a high-dimensional articulated system such as a full humanoid robot in a fraction of a millisecond~\cite{carpentier2019pinocchio}, if contacts do not exist. However, the bigger challenge for simulating legged locomotion is to include contact interactions. 

Early simulation environments sometimes modeled contact using springs and dampers, also known as penalty-based methods \cite{marhefka1996simulation,yamane2006stable,schaal2009sl}. While being able to simulate compliant contacts, these models by default result in large penetrations that do not accurately reflect reality. To address this requires increasing the stiffness of the system and therefore using small simulation time steps that result in slow simulations.
To avoid this \emph{stiff} numerical behavior, more recent simulators rely on rigid contact models, which can be either elastic or inelastic. In this approach, contact is handled via a complementarity condition together with friction cone constraints~\cite{smith2005open,todorov2012mujoco,coumans2016pybullet,lee2018dart,hwangbo2018per,makoviychuk2021isaac}. Access to fast simulation environments has been a key enabler for the success of deep reinforcement learning (DRL) for locomotion. GPU-friendly algorithms, e.g., \cite{makoviychuk2021isaac} has enabled fast training of complex robotics tasks on a single GPU \citet{kim2021survey}.
Importantly, it is equally relevant to be able to simulate the sensors of a legged robot, including robot-mounted cameras, and thus simulators have also seen advances in this regard.

\subsection{Control and Learning Algorithms}

Early methods to control quadruped robots commonly relied on the use of simplified template models \cite{blickhan1989spring,papadopoulos2000stable,kajita20013d,geyer2002natural} and simple-yet-effective heuristic-based strategies \cite{raibert1986legged,pratt1997virtual}. With the introduction of biologically-inspired approaches such as the central pattern generators (CPG) \cite{ijspeert2001connectionist}, a large body of work employed CPGs for generating oscillatory and cyclic behaviors for quadrupeds \cite{buchli2006finding,buchli2008self,ijspeert2008central,sprowitz2013towards,barasuol2013reactive}. CPG-like parameterizations are still adopted in several learning algorithms to expedite the training process \cite{miki2022learning,bellegarda2022cpg,ruppert2022learning,zhang2024learning} or to explain the motion of biological systems using robots \cite{shafiee2023puppeteer}. However, the enforced structure in these approaches limits the range of locomotion behaviors they can express, which can lead to sub-optimal performance. 

Currently, the two dominant approaches to controlling legged robots in multi-contact scenarios are optimal control (OC) and reinforcement learning (RL) methods.
OC uses a forward model of the system dynamics to help solve for a locally optimal control policy by minimizing a performance cost, typically over a finite future horizon~\cite{rakovic2018handbook}. In contrast, RL solves for a state-indexed optimal policy by maximizing the expected reward based on collected samples from rolling out a control policy \cite{sutton2018reinforcement}. The common ground between the two approaches is that they ultimately estimate an optimal policy for a desired task, as described via a cost or reward function. In recent years, both OC and RL have shown great success in the control of quadrupeds. 

Early work on the use of OC for quadrupedal locomotion cast the problem as a convex optimization problem, most typically as a quadratic program, either through using linear dynamics over a finite future time horizon \cite{kalakrishnan2011learning,park2015online,di2018dynamic,bledt2020extracting} or for the current instant in time, acting as an inverse dynamics controller \cite{buchli2009compliant,mistry2010inverse,righetti2013optimal,hutter2014quadrupedal,focchi2017high}. To enable planning and control for multi-contact behaviors through a holistic optimal control framework with contact, several formulations have been proposed, which are based on differential dynamic programming (DDP) with relaxed contact \cite{tassa2012synthesis}, contact-invariant optimization \cite{mordatch2012discovery}, contact-implicit optimization \cite{posa2014direct}, mixed integer convex optimization \cite{deits2014footstep,aceituno2017simultaneous}, and phase-based parameterization of the end-effector trajectories \cite{winkler2018gait}. While all of these approaches have shown impressive results in simulation, and some variants have become real-time capable for model predictive control (MPC) \cite{neunert2016fast,farshidian2017real,neunert2018whole}, their application in the real world has been limited. 
Most recent efforts to implement MPC on legged robots have focused on separating contact planning and whole-body motion generation and can achieve impressive behaviors on real hardware \cite{bledt2020extracting,li2021model,mastalli2023agile, grandia2023perceptive, meduri2023biconmp}.  

% These approaches consider a fixed contact plan and control the whole body motion for the given plan. To handle the complexity of the whole-body MPC for multi-limp systems, they use a hierarchy of MPCs to make the controller more reactive (adapting the location and time of contact) \cite{li2021model}, separate kinematics from dynamics \cite{meduri2023biconmp}, apply an instantaneous inverse dynamics to track the generated motions \cite{grandia2023perceptive}, leverage a memory of motions \cite{dantec2021whole} or offline trajectories \cite{mastalli2023agile} to warm-start MPC. 

While very successful, the use of optimal control and optimization-based control algorithms \cite{wensing2023optimization} comes with multiple challenges. 
Solving a high-dimensional non-convex optimization problem~\cite{wensing2023optimization} in real-time is computationally expensive \cite{sleiman2021unified,ponton2021efficient,mastalli2023agile,grandia2023perceptive,meduri2023biconmp}. 
Dealing with uncertainties, especially in contact interaction is difficult \cite{tassa2011stochastic,drnach2021robust,hammoud2021impedance,gazar2023multi}. 
The estimation and control problems are separated through the certainty equivalence principle, and the direct inclusion of different sensor modalities, such as vision, into control policy is prohibitively difficult. 
These issues motivated efforts towards alternative learning-based approaches for the control of legged robots.

% \changed{The Learning Locomotion project from DARPA drove the development of using learning-based approaches to solve rough terrain traversal problems for quadrupedal robots, e.g.,~\cite{pippine2011overview}. The works mainly focus on learning footstep planners, where potential footsteps are assigned scores using methods like apprentice learning~\cite{kolter2007hierarchical}, where human demonstrations are provided, or inverse learning~\cite{zucker2011optimization}, where humans assign footstep preferences between pairs of footstep as training signal.} 

The use of reinforcement learning for generating stable locomotion patterns has a history of more than two decades \cite{hornby2000evolving,kohl2004policy}. Early works commonly use hand-crafted policies with a few free parameters that are then tuned using RL on the hardware. Subsequent work employs the notion of the Poincare map to ensure the cyclic stability of the gaits \cite{tedrake2005learning,morimoto2005poincare}. These approaches have enabled a humanoid robot to walk \cite{morimoto2009nonparametric}, but their underlying function approximation has limited expressivity, which limits their application. Later reinforcement learning for quadrupeds mostly use the framework of stochastic optimal control, e.g., path integral policy improvement (PI$^2$) \cite{theodorou2010reinforcement,fankhauser2013reinforcement}.

The DARPA Learning Locomotion program led to a variety of learning-enabled solutions.
All the teams were provided with the Boston Dynamics LittleDog \cite{murphy2011littledog} such that they could focus on the software development.  
At the end of the program in 2009, four of six teams completed the challenge with satisfactory performance \cite{pippine2011overview}. 
To tackle a wide range of locomotion tasks on rough terrain, \cite{neuhaus2011comprehensive,shkolnik2011bounding} resorted to traditional planning and control algorithms. Other teams relied heavily on optimization with limited usage of learning techniques \cite{zucker2011optimization,zico2011stanford,kalakrishnan2011learning}. Notably, \cite{zico2011stanford} developed a hierarchical apprenticeship
learning framework \cite{kolter2007hierarchical} to approximate a cost map for path planning.  \cite{zucker2011optimization} used inverse optimal cost-preference learning to reduce the laborious cost-tuning procedure for footstep planning. The most extensive use of machine learning was demonstrated by \cite{kalakrishnan2011learning} where the authors developed a learning-from-demonstration framework to optimally select
foothold choices using terrain templates. Overall, despite the name of the program, learning techniques only constituted a small portion of the control software.
% In 2005, DARPA started the Learning Locomotion project that aimed to explore the use of machine learning
% techniques to accelerate the progress towards realizing autonomous legged systems \cite{pippine2011overview}. Six teams from American universities/institutions participated in this challenge and they were provided with the Boston Dynamics LittleDog \cite{murphy2011littledog} such that participants could focus on the software development. 
% A variety of increasingly complex environments were provided to the participants and their scanned versions were supplied to the control system in real-time.
% At the end of the program in 2009, four teams finished the challenge and demonstrated satisfactory performance by exceeding the metric speed, crossing large obstacles, and traversing highly rough terrains.

Thanks to the success of deep learning in the past decade \cite{lecun2015deep}, much of the focus of reinforcement learning has shifted to DRL.
Since DRL approaches commonly require a large number of samples (trial-and-error experience with the system being controlled)  to learn control policies, it is common practice to train the policies in simulation and then transfer them to the real world \cite{tan2018sim,hwangbo2019learning}. In contrast with MPC, DRL approaches make extensive use of offline simulation of the world to learn a policy. Once learned, the online computation for the deployed policy consists of a simple forward pass through the neural network (often referred to as `inference') at each control time step, which is fast and efficient to compute. Furthermore, while training policies using DRL, robustness to many types of uncertainty can be achieved by adding noise during training to the relevant robot parameters and environments \cite{lee2020learning,bogdanovic2022model}. Finally, it is simple to directly incorporate arbitrary sensory data as input to the policy, including high-dimensional modalities such as vision. This largely eliminates the need for a separate state estimation module \cite{miki2022learning,agarwal2023legged}, as state estimation becomes an implicit part of the control policy. These unique features have led to the success of DRL in controlling quadruped robots on various terrains and under a wide range of conditions.

\subsection{Related Survey Papers}

Several existing survey papers have some overlap with the scope of our survey. The most relevant is \cite{zhang2022deepreinforcement} which briefly surveys the use of DRL for quadrupedal locomotion. Another recent survey paper \cite{bao2024deep} reviews the progress of DRL for bipedal locomotion. In comparison, this paper aims to provide a more comprehensive overview of the methods and complexities in this area. We cover not only DRL but also other types of learning for quadrupedal locomotion, such as imitation-based learning methods. We provide a historical perspective and further describe how learning-based approaches complement the rich literature on model-based control methods for legged locomotion. We further discuss ongoing advances and open directions. 
Complementary to our survey, \cite{wensing2023optimization} focuses on optimal control methods for locomotion, while \cite{darvish2023teleoperation} reviews works on the teleoperation of humanoid robots. The survey provided in \cite{ibarz2021train} has a focus on DRL for manipulation. While the focus in this survey is on quadrupeds and bipeds, the bulk of the foundations and algorithms are also relevant for robots with an arbitrary number of legs, from one-legged \cite{bogdanovic2020learning,bussola2024efficient} to six \cite{schilling2021adaptive} and eight-legged robots \cite{li2020learning,li2021planning}.

Advances in simulators and simulation environments have been key to enabling deep learning for robotics. Numerous high-performance open-source simulators are now available and are accelerating progress, e.g., \cite{coumans2016pybullet,makoviychuk2021isaac,mujoco2023jax}. 
Multiple recent works have extensively reviewed the state of the art in this area and provide detailed comparisons, e.g.,~\cite{collins2021review,kim2021survey,liu2021role}.

\section{Theoretical background}\label{sec:background}
% \subsection{Model-based Control}
% \majid{please check the "control algorithms" in the introduction. There I covered model-based optimal control. I think we can skip here talking about model-based approaches (I think we can still talk about model-based RL)}
% - Model-based control is to illustrate the given locomotion problem using simplified model and find optimal actions via optimization algorithms.
% - Although it is not our main focus, it might be great to discuss this briefly to highlight the similarity and difference.
% - In this philosophy, researchers capture the dynamics using various simplified models
% - Then QP is formulated; which looks like:
% - Model deviates; therefore we need a replan at high-frequency
% - For more details, please refer to the survey papers (there must be some...)

% - Model-based control has been effective for legged locomotion for more than great decades.
% - But it is based on the assumptions, which sometimes limits its 
% - Multi-staged formulation also requires large computational costs
% - Learning splits this into: a substantial training and a brief testing.

\begin{figure*}
    \centering
    \includegraphics[width=\linewidth]{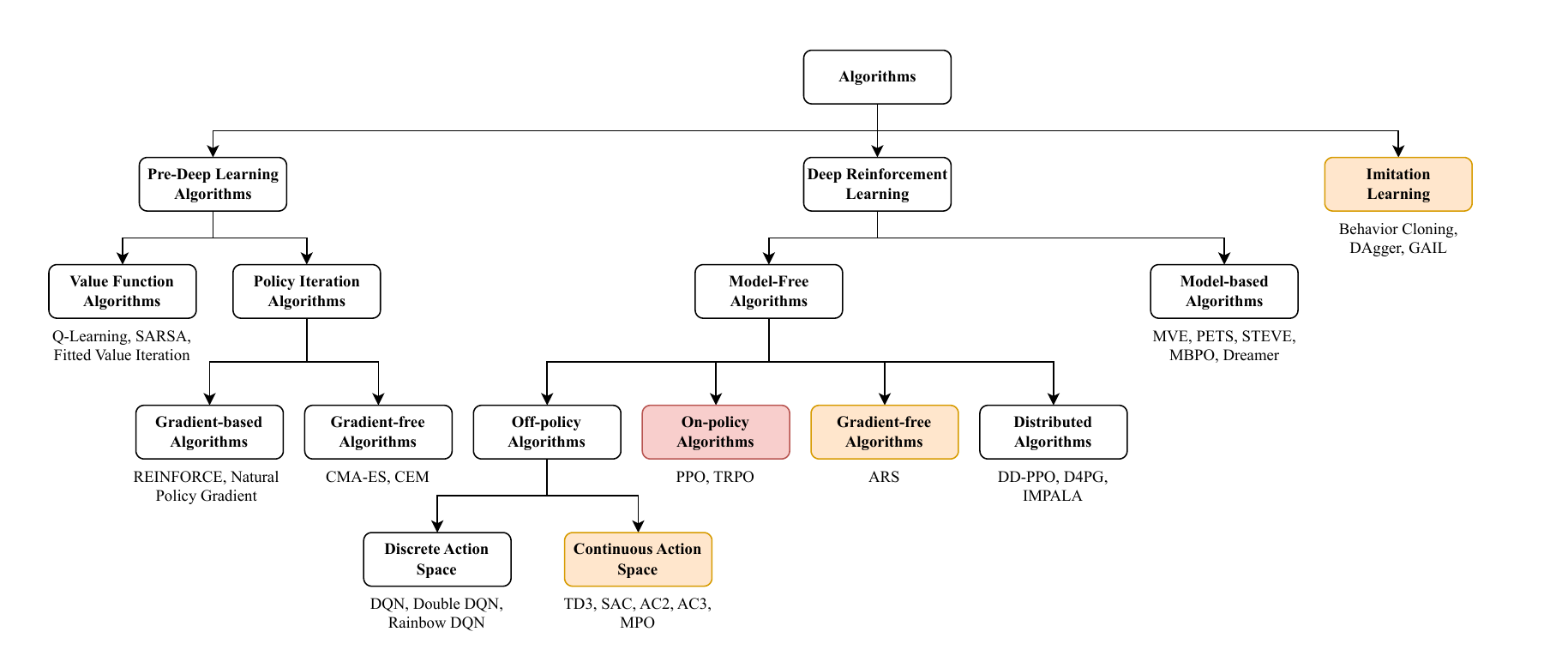}\vspace{-5mm}
    \caption{Taxonomy of reinforcement learning algorithms. Popular algorithms are highlighted in color. On-policy algorithms (red), such as TRPO or PPO, have been the most frequent choice for legged locomotion. Other off-policy, gradient-free, or imitation learning algorithms (orange) are also selected for sample efficiency or to reproduce styles from example data. For further details, please refer to the background section.}
    \label{fig:taxonomy}
\end{figure*}

\subsection{Markov Decision Process and Reinforcement Learning}\label{subsec:MDP}
For many decades, machine learning (ML) has demonstrated promise for solving complex problems across a wide range of domains including computer vision, natural language processing, finance, environmental science, and many more. In robotics, RL~\cite{sutton2018reinforcement} stands out as a popular choice because the control of robots can naturally be viewed as a sequential decision-making problem. RL is one of the ML paradigms that seeks to develop an intelligent agent in dynamic environments to maximize cumulative rewards. 

Typically, RL models the control problem using the Markov decision processes (MDP), which is defined as a tuple $(\sspace, \aspace, p, r, \gamma)$ of the state space $\mathcal{S}$, action space $\mathcal{A}$, deterministic $\mathbf{s}_{t+1} = f(\mathbf{s}_t, \mathbf{a}_t)$ or stochastic transition function $p(\stp|\st, \at)$, reward function $r(\st, \at, \stp)$, and a discount factor $\gamma$.
By executing a policy $\pi(\at|\st)$ in a stochastic environment, we can generate a trajectory of state and actions $\tau = (\state_0,\action_0, \state_1, \action_1, \ldots)$.
We denote the trajectory distribution induced by $\pi$ as $\rho_\pi(\tau)=p(\mathbf{s}_{0})\prod_t\pi_t(\mathbf{a}_t|\mathbf{s}_t)p(\mathbf{s}_{t+1}|\mathbf{s}_t,\mathbf{a}_t)$. Our goal is to find the optimal policy $\pi^*$ that maximizes the sum of expected returns:
\begin{equation}
    J(\pi) = \E{\tau\sim \rho_\pi}{\sum_{t = 0}^{T} r(\st, \at)}.
    \label{eq:rl_objective}
\end{equation}

In basic movement scenarios that involve navigation over a flat ground plane, the MDP state is given by the state of the robot alone, i.e., the information needed to describe the positions and velocities of all the links of the robot. In more interesting and complex scenarios, the MDP state should encapsulate both the state of the robot and the state of the surrounding environment. 
In practice, many variations of MDPs are commonly used in order to accommodate diverse scenarios. For example, partially-observable MDPs (PoMDPs) capture the need to make control decisions using incomplete observations instead of the full MDP state, as defined by the observation space $\ospace$ and the observation function $o(\mathbf{o}_t|\mathbf{s}_t)$. The PoMDP formulation is common in robotics, where a robot camera can only observe part of the world at any given time, for example. The available observations are sometimes directly referred as the state, $\mathbf{s}_t$, which is a slight abuse of notation that allows policies to always be described as performing a state-to-action mapping, even when the actual mapping being used is from observations-to-actions.

Researchers have developed various RL algorithms to address MDP problems. Early algorithms can be broadly categorized into value-function and policy iteration methods. Value-function methods, exemplified by Q-learning~\cite{watkins1992q}, SARSA~\cite{rummery1994line}, or (Fitted) Value Iteration~\cite{bellman1966dynamic,boyan1994generalization}, aim to estimate the expected values of states or state-action pairs, which implicitly define the optimal policy. Conversely, policy iteration methods, whether gradient-based~\cite{howard1960dynamic} or gradient-free~\cite{hansen2003reducing}, frame the given MDP as an optimization problem and seek to find the best parameters from a policy standpoint. These algorithms have found successful applications in diverse robotics problems, including legged locomotion.

Despite promising results, these early algorithms encounter challenges in solving complex, large-scale control problems, mainly due to scalability and convergence issues. Nonetheless, the established theoretical advancements demonstrate significant performance enhancements when combined with deep learning, which opens a new era of research.

\subsection{Deep Reinforcement Learning}
Deep learning~\cite{lecun2015deep,goodfellow2016deep} stands out as one of the most significant breakthroughs in ML. Artificial neural networks, composed of layers of neurons, are recognized as universal function approximators that can learn any multi-variate functions~\cite{hornik1991approximation,cybenko1989approximation}. Various researchers have demonstrated that neural networks can effectively solve complex regression problems by processing vast amounts of data if they are provided with appropriate network architectures and learning frameworks.

The pioneering work of \cite{mnih2015human} demonstrated that RL can also be combined with deep learning to achieve human-level performance for Atari games. This achievement was made possible by introducing replay buffers and target networks to stabilize learning. Following Deep Q Network (DQN), several innovations have emerged to improve convergence and sample efficiency~\cite{wang2016dueling,van2016deep,hessel2018rainbow}. While many algorithms addressed simple discrete actions, robotic control usually requires continuous action spaces to generate motor commands. Deep deterministic policy gradient (DDPG)~\cite{lillicrap2015continuous} was one of the first algorithms supporting continuous actions, followed by numerous other RL algorithms and variations, e.g.,~\cite{fujimoto2018addressing,haarnoja2018soft,mnih2016asynchronous,abdolmaleki2018maximum,schulman2015trust,schulman2017proximal}. 
Model-based RL holds the premise of achieving the best sample efficiency by utilizing the learned world model of the given MDP either in state space~\cite{nagabandi2018neural,feinberg2018model,buckman2018sample,chua2018deep,janner2019trust} or image space~\cite{ha2018world,hafner2019learning,hafner2019dream}. Nevertheless, recent model-free approaches~\cite{chen2021randomized,hiraoka2021dropout} have also shown competitive sample efficiency. Besides sample efficiency, there have been numerous efforts to improve the scalability of RL algorithms for both gradient-based methods~\cite{espeholt2018impala,barth2018distributed,wijmans2019dd} and gradient-free approaches~\cite{mania2018simple}. 

In the domain of learning-based locomotion, on-policy model-free algorithms like trust region policy optimization (TRPO)~\cite{schulman2015trust} and proximal policy optimization (PPO)~\cite{schulman2017proximal} are often preferred choices. This preference arises from a pursuit of optimal-and-robust performance while being less concerned with sample efficiency. PPO has been a particularly popular choice in the legged locomotion community due to its excellent convergence and adaptability to diverse policy architectures. 
Table~\ref{tab:usecases} provides a detailed taxonomy of the most common RL algorithms, and identifies those most commonly used for learning legged locomotion.

It is important here to clarify one common misconception when the term \emph{model-free} is used in the context of DRL in robotics. While none of the model-free DRL algorithms need a model of physics of the world \emph{per se}, in practice these policies are trained in simulation environments that are physics models. This means that application of model-free DRL methods relies heavily on models (simulation environments) that are developed using first principles. Hence, the term \emph{model-free} in the context of DRL in robotics should be taken with a grain of salt.

\begin{table*}[h]
    \centering
    \caption{Common algorithms and use-cases for legged locomotion. A popular option has been on-policy RL, such as TRPO or PPO, due to their convergence to the best performing policies. However, other algorithms have been used for other reasons, such as better explotation, sample efficiency, or scalability.}
    \begin{tabular}{|c|c|c|p{0.5\textwidth}|}
        \hline
        \multicolumn{3}{|c|}{} & Example Papers \\
        \hline
        \multirow{5}{*}{\makecell{Reinforcement\\Learning}} & \multirow{2}{*}{\makecell{Off-policy\\Learning}} & DDPG & \cite{bogdanovic2020learning,huang2022reward} \\ \cline{3-4}
         &  & SAC & \cite{haarnoja2018learning,ha2020learning,smith2023learning} \\ \cline{2-4}
         & \multirow{2}{*}{\makecell{On-policy\\Learning}} & TRPO & \cite{hwangbo2019learning,lee2020learning,yang2022safe}  \\ \cline{3-4}
         &  & PPO & \cite{iscen2018policies,tan2018sim,peng2020learning,kumar2021rma,margolis2021learning,rudin2022learning,xie2022glide,miki2022learning,zhuang2023robot,agarwal2023legged,liu2024visual} \\ \cline{2-4}
         & Model-based & Dreamer &\cite{wu2023daydreamer} \\ \cline{1-4}
         & Gradient-free & ARS &\cite{yu2020learning,yu2021visual} \\ \cline{1-4}
         
         \multirow{2}{*}{\makecell{Imitation\\Learning}} &  \multicolumn{2}{c|}{Behavior Cloning}  & \cite{lee2020learning,kumar2021rma,liu2024visual,reske2021imitation} \\ \cline{2-4}
          &  \multicolumn{2}{c|}{GAIL}  & \cite{escontrela2022adversarial} \\ 
        \hline
    \end{tabular}
    \label{tab:usecases}
\end{table*}

\subsection{Behavior Cloning and Imitation Learning}
As mentioned, the main goal of RL is to find control policies that maximize the cumulative reward, in a model-based
or model-free fashion. Either way, we assume that the reward function is given to us and the algorithm collects data to learn a model (model-based), or a value/policy function (model-free). While this has shown to be very promising in practice, it requires an immense amount of reward engineering, in particular for problems with sparse rewards. 

Learning to imitate actions from an expert human demonstration or an existing expert policy~\cite{pomerleau1988alvinn} can be seen as a remedy to this problem. However, naively mimicking the actions of the expert often fails due to the compounding nature of errors in control problems and the resulting distribution mismatches between the states visited during training and those seen at runtime. To mitigate this issue, researchers have explored alternative approaches for developing robust policies, such as dataset aggregation (DAgger)~\cite{ross2011reduction} or generative adversarial imitation learning (GAIL)~\cite{ho2016generative}. One notable usage of IL in learning-based locomotion is privileged learning~\cite{chen2020learning}, which first learns a capable teacher policy with ground-truth information and copies the behaviors into a student policy with a realistic sensor configuration. Given its importance, we will revisit this particular strategy later in Sec. \ref{subsec:privileged_learning}.

\section{Components of MDP for Locomotion}
As mentioned in Sec. \ref{sec:background}, RL algorithms typically frame sequential decision-making problems as MDPs. In this section, we will provide a brief summary of common practices in MDP formulation for legged robot control problems, including dynamics, state/observation, reward, and action space.

\subsection{Dynamics}
A legged robot can be described as a (typically deterministic) dynamical system, with the equations of motion given by $\dot{\mathbf{s}} = f(\mathbf{s}, \mathbf{a})$, where $\mathbf{s}$ is the state of the robot and the environment, and $\mathbf{a}$ is the action taken by the robot. To formulate it as an MDP, we discretize the dynamics with respect to time, $\mathbf{s}_{t+1} = \mathbf{s}_t + f(\mathbf{s}_t, \mathbf{a}_t) dt$, with $dt$ being the discretization step. 
In MDP notation, this is modeled using a stochastic transition function $p(\stp|\st, \at)$
having the probability mass lumped at $\stp$.
The state-transition data tuples are generated from the dynamics, $(\st,\at,\mathbf{r}_{t},\stp)$, and used for the RL algorithms.

\subsubsection{Simulators}
% \begin{table*}[]
%     \centering
%     \caption{simulation environments used for learning quadrupedal locomotion}
%     \begin{tabular}{|c|c|}
%     \hline
%         Bullet & \cite{tan2018sim,peng2020learning,bogdanovic2022model} \\ \hline
%         Mujoco &  \cite{brakel2022learning,bohez2022imitate}\\ \hline
%         Isaac &  \cite{rudin2022learning,zhuang2023robot,nahrendra2023dreamwaq}\\ \hline
%         RaiSim & \cite{hwangbo2019learning,lee2020learning,gangapurwala2021real,choi2023learning} \\ \hline
%         Webots &  \cite{tan2021hierarchical}\\
%     \hline
%     \end{tabular}
%     \label{tab:simulations}
% \end{table*}

Since DRL requires a large number of samples to train a policy, it is common practice to learn policies in simulation and then directly deploy them in the real world. As noted in Sec. \ref{subsec:simulators}, fast and accurate simulators have played a critical role in recent advances in learning locomotion behaviors.

\begin{figure}
    \centering
    \includegraphics[trim={1cm 0 0 0},clip,width=.99\linewidth]{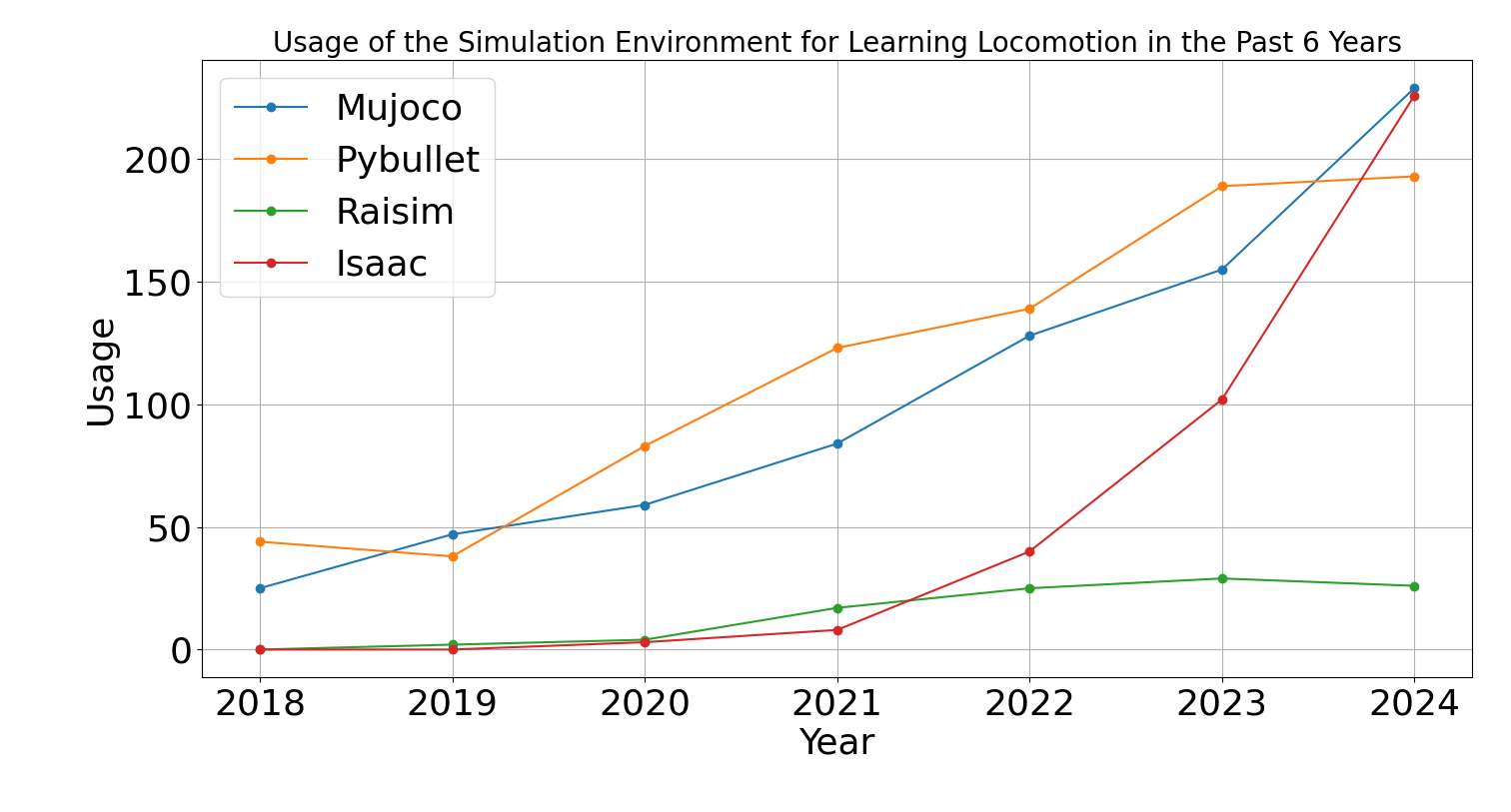}\vspace{-5mm}
    \caption{A rough estimation of the usage of various simulators for locomotion learning. The numbers are collected using a Google Scholar search with the following keywords "simulator name"  "legged" "robot"  "hardware" and "learning".}
    \label{fig:sim}
\end{figure}

State-of-the-art simulators used in robotics commonly rely on a rigid contact (elastic or inelastic) assumption. They model contact using a complementarity condition together with friction cone constraints (linear or nonlinear) \cite{todorov2012mujoco,coumans2016pybullet,hwangbo2018per,makoviychuk2021isaac}. Also, there has been a tremendous effort towards making these simulation environments differentiable \cite{geilinger2020add,le2021differentiable,howell2022dojo} which would make it possible to perform efficient system identification through contact or to directly update the parameters of a policy via backpropagation through time. Multiple recent works have extensively compared the simulation environments and we refer the readers to those works for a complete survey \cite{collins2021review,kim2021survey,liu2021role} and to \cite{lidec2023contact} for an extensive survey of the multitude of methods used by simulators to solve for (rigid) contact in robotics. The GPU-friendly nature of Isaac Sim \cite{makoviychuk2021isaac} allows for collecting state-transition data at a rate of nearly one million per second, making it especially interesting for DRL \cite{kim2021survey}. Relatedly, the new version of MuJoCo includes MuJoCo XLA (MJX) which makes it GPU-compatible via the JAX framework \cite{mujoco2023jax}. 
The trend of recent simulator usage for learning-based locomotion is illustrated in Figure~\ref{fig:sim}, which captures early-developed communities for PyBullet and MuJoCo and also highlights the recent rise of Isaac Sim.

Relying on a rigid contact model that disallows any compliance or penetration is a limitation when dealing with tasks where interacting with a compliant environment is essential. Recently, researchers in both fields of locomotion and in-hand manipulation have looked into more advanced contact models to better capture the complexity of compliant interactions \cite{khadiv2019rigid,masterjohn2022velocity,choi2023learning}. For instance, \cite{khadiv2019rigid} proposes a nonlinear compliant contact model that can handle both rigid and soft interaction scenarios. In \cite{masterjohn2022velocity}, the idea of pressure field contact patches is re-visited to foster their use with available velocity-level time steppers which enable simulation of compliant interactions with real-time rates. In \cite{choi2023learning}, on the other hand, an analytical nonlinear point contact model is used. To match the behavior of the model with the real-world data, the authors randomized the stiffness and damping ratio of the terrain such that the distribution of the simulation results matched the experimental distributions. To enable interaction with a wide range of environments and objects, a crucial component for learning-based algorithms will be reliable and fast simulation environments that can model a wide range of physical interactions.

\subsubsection{Real robot}
Another possible choice for generating motion data is to directly use the real robot to record the reward~\cite{saggar2007autonomous} or the resulting transitions $\{\mathbf{s}_t,\mathbf{a}_t, \mathbf{s}_{t+1}\}$~\cite{haarnoja2018learning,choi2019trajectory,yang2020data}. This approach can potentially guarantee performance in the real-world environment by avoiding the need to build a model of the system and ensuring the fidelity of the data being used to train our ML model. However, collecting such data can be costly, which greatly limits its scalability. In addition, safety becomes a more important issue, especially for legged robots, because their underactuated base makes them vulnerable to falls and reliable mechanisms do not exist for resetting to a known initial state to begin every new episode. Therefore, additional mechanisms are required to mitigate the safety issues, such as safety-aware RL~\cite{ha2020learning,yang2022safe}, automatic resetting~\cite{luck2017lab,ha2018automated}, failure recovery~\cite{smith2022walk}, or simulation-based pre-training~\cite{smith2023learning}.
% A straightforward way to generate dynamic data is to directly feed action input to the robot we want to control and record the resulting transition $\{\bf{s}_t,\bf{a}_t, \bf{s}_{t+1}\}$. This avoids the need to build a model of the system and ensures the fidelity of the data being used to train our machine learning model. However, collecting such data can be costly, as robots can be broken by unsafe actions, especially for legged robots, which are inherently unstable dynamical systems. Additional machinery like learning the set of safe actions, and learning to recover from failed states, e.g., getting back up after a fall is necessary.

\subsection{Observation}
One crucial aspect of MDPs is the observations and how to infer the states from them. The observation space is comprised of noisy sensory measurements that give information about the state of the robot and the environment. In the following, we categorize the body of works in the field of locomotion based on the types of observations they use for learning policies.
% Relying on the certainty equivalence principle, one can infer the states of the system form the measurements independent of the task and control policy. In this case, the control policy assumes access to the true states of the system and design the control policy as if it has access to the true state of the system. Another approach is to 

% \majid{We should think about how planning/hierarchy fit to the story}
\subsubsection{Proprioception}
The set of sensors that provide information about the internal state of the robot is called proprioception, like IMU, joint encoders, or contact sensors.
In legged robotics, raw measurements from these sensors are typically not directly used due to potential noise and inaccuracies. Instead, a state estimator is typically used to estimate crucial states of the robot such as base pose, base twist, joint position, and velocity.
Such states are often referred to as \emph{proprioceptive states} in DRL works, including~\cite{lee2020learning,nahrendra2023dreamwaq,yang2021learning,yu2023identifying}.

Recently, \cite{ji2022concurrent} and \cite{lee2024AOW} have used sequences of IMU measurements and joint encoder data directly, instead of relying on pre-estimated velocities and poses. Specifically, \cite{ji2022concurrent} implemented an RNN-based state estimator, diverging from traditional model-based estimators, to directly compute the base velocity and contact states. On the other hand, \cite{lee2024AOW} employed a privileged training strategy, effectively utilizing the raw sensor data sequences to enhance the robustness.

The emerging research in legged robotics highlights the pivotal roles of joint positions, base pose (often represented by the gravity vector), and base linear and angular velocities in developing robust locomotion skills~\cite{yu2023identifying}. A significant body of work, including \cite{lee2020learning}, \cite{ji2022concurrent}, and \cite{haarnoja2018learning} emphasize the importance of leveraging a history of proprioceptive states and joint command information. This approach addresses the non-Markovian characteristics of robotic systems that stem from hardware latencies and partial observations.
Using only the current state is typically insufficient due to state estimation errors and partial observability issues in the real world; therefore, a short history buffer is almost always employed.
In particular, historical data of proprioceptive states provide critical insights into foot-terrain interactions and external disturbances, as demonstrated by \cite{lee2020learning}, \cite{ji2022concurrent}, and \cite{li2024reinforcement}. Also, \cite{peng2018sim} examines the benefits of including history inputs, demonstrating that under dynamics randomization, a policy with memory outperforms the ones without.

% Recent works say
%  that  joint positions, base pose (often represented by gravity vector) and base linear and angular velocities play the crucial role in achieving robust locomotion skills. 
% Many works employ a history of such proprioceptive states and joint command information (e.g., position tracking errors~\cite{hwangbo2019learning} or action history~\cite{ji2022concurrent,haarnoja2018learning}) to address the non-Markovian nature of robotic systems due to latencies in the hardware and partial observation.
% The history  conveys information about foot-terrain interaction and external disturbances as shown by \cite{lee2020learning} and \cite{ji2022concurrent}. 
% By using kinematic information collected over time, robots are better equipped to make more informed decisions, significantly improving their ability to operate effectively in complex, real-world scenarios. 

% In the community, the term ``blind locomotion'' often refers to legged robot control only with proprioception, which can still overcome quite challenging terrains. 

\subsubsection{Exteroception}
The sensory layer of the robot that provides information about the surrounding environments is normally referred to as exteroception. This information is in particular vital when the robot moves on non-flat environments.
Conventional approaches in legged robotics often rely on explicit mapping techniques to pre-process the measurements before they are used by the controllers. 
For example, methods like elevation mapping~\cite{miki2022elevation} or voxel mapping~\cite{oleynikova2017voxblox, besselmann2021vdb} are commonly employed.
Early works on perceptive locomotion~\cite{miki2022learning, xie2022glide} adopted elevation mapping, where height values around the robot were sampled and used as observation to the policy. 

Recent works removed such explicit mapping and instead used raw sensory readings directly as input to the policy, such as depth images or point clouds.
This shift is driven by the necessity to effectively handle highly dynamic situations such as parkour~\cite{zhuang2023robot} or obstacle avoidance~\cite{yang2021learning}, and tasks requiring high-resolution perception, such as stepping stone scenarios~\cite{duan2022learning,zhang2023learning}.
This approach not only facilitates more responsive and accurate locomotion in complex environments, but also reduces the likelihood of errors and inaccuracies that often arise from mapping failures.

RGB data can also be used for a more evolved perception of complex scenes that go beyond mere geometric information, enabling sidewalk navigation and obstacle avoidance~\cite{sorokin2022learning}.
This modality includes a richer context, recognizing and responding to elements that cannot be captured through geometry alone, such as textures and colors. Semantics of the scenes can also be learned for more efficient navigation. For instance, \cite{yang2023learning} utilize semantic information from images to adapt the locomotion gaits and speed of a quadruped to handle different terrains.
\cite{margolis2023learning} further augment RGB data with proprioceptive data for better traversability estimation.

Instead of feeding the sensory values directly into the policy, a compressed representation of them can be learned. For example, \cite{hoeller2021learning, yang2023neural} use unsupervised learning to obtain a latent space that can compress and reconstruct the original camera images, enabling learning policies that can navigate complex terrains automatically.

% \subsubsection{Learned State Representations}
% Some works leverage learned state representations, which are latent spaces of encoders or world models, as observations ~\cite{hoeller2021learning}.

\subsubsection{Task-related inputs (goals)} 
Apart from proprioception and exteroception, other information specific to the robot's task, including command inputs like velocity and pose, or more complex data representations such as learned task embeddings can be included as input to the policy. Note that these inputs can be seen more as goals than observations, but to not overload the sections in the paper, we cover them under observation.

Velocity commands are frequently used to steer locomotion. Pose commands, specifying particular position or orientation targets, are also employed~\cite{rudin2022advanced}.
In addition to direct command inputs, some approaches involve the use of learned task embeddings, which could be a latent representation of a specific reference motion~\cite{peng2022ase} or any latent space that can direct the low-level behavior~\cite{haarnoja2018latent}.
For robotic systems employing structured action spaces, such as those based on CPGs or trajectory parameters, task-related information can also include specific details about the desired phase or trajectory patterns~\cite{iscen2018policies, lee2020learning}.

Some methods consider future reference trajectories as input to the policies. For example, \cite{ma2022combining} uses planned end-effector trajectories of a manipulator as additional input for a quadrupedal robot with an arm, allowing the policy to adjust whole-body motions in anticipation of the arm's movements.
\cite{gangapurwala2022rloc} and \cite{jenelten2024dtc} provide planned footholds and trajectories as references to train a tracking controller. This approach, where model-based controllers supply motion trajectories as references to the policy, enables more versatile solutions to complex tasks like navigating stepping stones. However, it also increases complexity and engineering effort due to the need for additional planning and coordination between controllers.
Consequently, many recent works on dynamic locomotion prefer to learn a single independent control policy without any reference trajectory \cite{hoeller2023anymal}, \cite{zhuang2023robot}, \cite{cheng2023extreme}.

\subsection{Reward}\label{subsec:reward}

The reward function plays a crucial role in delivering the desired behavior by the RL algorithm, as it defines the objective of the agent.
A common approach is to formulate the reward function as a linear combination of various reward and penalty terms, denoted as $r(\mathbf{s}_{t}, \mathbf{s}_{t+1}, \mathbf{s}_{t}) = \sum_{i}c_i r_i(\mathbf{s}_{t}, \mathbf{s}_{t+1}, \mathbf{a}_{t})$, while it is also possible to use multiplicative combination~\cite{kim2022humanconquad}.
There are various methods to define each reward component.

\subsubsection{Manual reward shaping}
Manual reward shaping involves defining each reward component $r_i$ and tuning each weight $c_i$ by the engineer (hence a.k.a. reward engineering). Some examples of these components are velocity tracking, pose tracking, and other regularization terms. Also, normally the physical constraints of the system are added as a cost to this function, e.g., limiting the magnitude of joint velocity, acceleration, and base pose during locomotion.

There is no general set of rules one can follow to define the components of the reward function.
One common practice is to use bounded functions for stable training. This is often done by simply clipping or applying exponential kernels. 
For instance, exponential functions such as $exp(-c \lvert\lvert e\rvert\rvert^2)$ or $exp(-c \lvert\lvert e\rvert\rvert)$, with $e$ representing the error term and $c$ being the shaping coefficient, are frequently employed for tracking tasks~\cite{lee2020learning,rudin2022learning, duan2022learning}. A set of commonly-used physical quantities used to define reward components are summarized in Table \ref{tab:reward}. 
To make the shaping of the reward terms easier, recent works have looked into alternatives such as potential-based rewards \cite{jeon2023benchmarking}

% Additionally, a curriculum is frequently introduced to the regularization reward functions~\cite{hwangbo2019learning}. This involves gradually increasing the scale of such terms. This relaxation improves exploration at the early phase of the training and results in a higher final performance.

\begin{table}[]
    \centering
%     \caption{Typical rewards used for training quadruped locomotion. 
% $*$ denotes for target quantity
% $\phi(x) := exp(- k \lvert \lvert x \rvert \rvert ^2)$ 
% $k$ is a scaling factor. 
% $e_z^b$ is the world z axis in base frame.
% $q$ joint angle
% $\tau$ joint torque
% $a_t$ action.
% Reward functions can be similarly defined using different norms or element wise scales.
% % $R_{bw} $ is the orientation of the base frame with respect to the world frame
% }
\caption{Common physical quantities used to define rewards for training quadruped locomotion. 
Various norms and kernel functions are then applied to reward or regularize these terms. 
An asterisk ($*$) indicates a target quantity. 
% \changed{\st{
% The function $\phi(x) := \exp(-k \lvert \lvert x \rvert \rvert^2)$, where $k$ is a scaling factor, is used to define bounded rewards. 
% }}
$e_z^b$ represents the z-axis of the world in the base frame, $q$ denotes the joint angle, $\tau$ represents the joint torque, and $a_t$ is the action. Reward functions can be similarly defined using different norms or element-wise operations.
}

    % \cite{nahrendra2023dreamwaq,rudin2022learning}}
    \begin{tabular}{c|c}
         Horizontal velocity error & $ v_{xy}^{*}-v_{xy}$ \\ \hline
         Yaw rate error & $\omega_{z}^{*}-\omega_{z}$ \\ \hline
         Base vertical velocity & $v_z$ \\ \hline
         Roll and pitch rate & $ \omega_{xy}$ \\ \hline
         $z$-axis deviation & $\lvert \lvert [0, 0, 1]^T - e_z^b \rvert \rvert $ \\ \hline
         Joint velocities & $\sum_{i \in \text{joints}}\| \dot{q_i}\|$ \\ \hline
         Joint accelerations & $\sum_{i \in \text{joints}}\| \ddot{q_i}\|$ \\ \hline
         Joint torques & $\sum_{i \in \text{joints}}\| {\tau_i}\|$ \\ \hline
         Joint mechanical power & $\sum_{i \in \text{joints}} {\tau_i} * \dot{q_i}$ \\ \hline
         Action rate & $ \|\mathbf{a}_t - \mathbf{a}_{t-1}\|$ \\ \hline
         Action smoothness & $\|\mathbf{a}_t - 2\mathbf{a}_{t-1} + \mathbf{a}_{t-2}\|$
    \end{tabular}
    \label{tab:reward}
\end{table}

\subsubsection{Imitation reward}
We usually have a rough idea of how a legged robot should move due to the biologically inspired nature of these robots. These behaviors can be obtained from motion capture data of humans and animals (or online videos). Reward signals can be designed using this information, which greatly reduces the engineering effort required.
% An alternative approach is to replace reward components with an imitation objective if motion capture data or demonstration data is available. This approach greatly reduces the engineering effort required to define and tune each reward component manually.

There exist many works~\cite{peng2018deepmimic,peng2020learning,han2023lifelike,yang2023generalized} that leverage dog motion capture data to enable quadrupedal robots to learn animal-like motions, which is often approached by traditional model-based control approaches~\cite{kang2021animal,li2023fastmimic}. In this motion imitation problem, the reward is designed to simply imitate the given reference trajectory, which can be obtained by either retargeting motion capture data of a real canine~\cite{peng2020learning,klipfel2023learning}, model-based controllers~\cite{reske2021imitation,fuchioka2023opt}, or solving trajectory optimization~\cite{bogdanovic2022model,kim2023armp}. 

Instead of simply tracking the given reference motion, a set of motion clips can be used to define the style of legged robot movements, such as gait frequency and base motions. Generative adversarial imitation learning (GAIL)~\cite{ho2016generative} simultaneously trains a discriminator along with a policy, which distinguishes whether the motion is from the existing database or generated by a policy. Then, the learned discriminator can serve as a generic motion prior to realistic or stylistic movements, which can be co-optimized with downstream task rewards. The concept of adversarial motion priors (AMP) as a substitute for complex reward terms has also been used in multiple recent works~\cite{escontrela2022adversarial,wu2023learning,vollenweider2023advanced,li2023learning,li2023crossloco}. For periodic or quasi-periodic motions, one can also enforce more structure to extract spatio-temporal relationships in demonstrations \cite{li2024fld}.

% \majid{Adversarial motion priors papers to be discussed here \cite{escontrela2022adversarial,wu2023learning,vollenweider2023advanced,li2023learning}}

% This is also closely related to imitating trajectories generated by model-based controllers, such as MPC, to produce smooth and near-optimal motions with RL-based controllers \cite{reske2021imitation,kang2021animal,li2023fastmimic}.
% For instance, \cite{han2023lifelike} used dog motion capture data to enable quadrupedal robots to learn animal-like motions. 
% Previsouly, some works imitate trajectories generated by model-based controllers, such as MPC, to produce smooth and near-optimal motions with RL-based controllers \cite{reske2021imitation}.
% This is implemented by training the agent to imitate expert demonstrations, either by explicitly mimicking the joint positions~\cite{peng2018deepmimic} or through inverse reinforcement learning (IRL) approaches with learned discriminators~\cite{ho2016generative}.
% The IRL approach implicitly captures motion styles such as gait frequency and base motions.

\subsection{Action Space}
Action space plays an important role in the performance of learned controllers for robotics, especially in the case of RL where exploration happens in the action space, the appropriate choice of action space can greatly improve exploration efficiency.

\subsubsection{Low-Level Joint Commands}
% directly command joint actuator commands such as joint position, velocity, or torque.
% Someone~\cite{} benchmarked different formulations and joint position target is the most commonly used

Most works in quadrupedal learning use joint target position as the action space  (PD policy). For each joint of the robot and a given action value $a(s)$ for that joint, the motor will attempt to generate a torque of $\tau = k_p (a(s) - \theta) - k_d \dot{\theta}$ to move the joint ($k_p$ and $k_d$ are the gains, while $\theta$ and $\dot{\theta}$ are the measured joint angles and velocities). In addition to its practicality, another reason this is the default choice may be attributed to \cite{peng2017learning}, where they show using joint position target as action space performs the best for physics-based character animation tasks, compared to other choices such as torque and velocity. Also, \cite{bogdanovic2020learning} systematically showed on a one-legged hopping task that varying the gains of the controller as a function of states can improve the performance, when contact is highly uncertain.

More recently, \cite{chen2023learning, kim2023torque} also demonstrates successful learned quadrupedal and bipedal behaviors without imposing any structure in the action space and directly outputting torque (torque policy). The main benefit of this approach is that the policy function is not limited by the choice of the imposed structure as in the PD policy case. However, in this case, one needs to increase the frequency of policy evaluation at run-time (e.g., 1 KHz), whereas the PD policy can be updated much slower (e.g., 50 Hz) with torque still being produced at 1 KHz. Furthermore, it is argued in \cite{bogdanovic2020learning} that the sim-to-real transfer for the PD policy is easier. It is important to note that, as it is also explained in \cite{hwangbo2019learning}, the PD policy is different from position control traditionally used in robotics. For position control, the controller tracks a \emph{desired time-indexed trajectory} with large gains. However, no desired trajectory is tracked by the PD policy (note that the desired velocity is zero) and the target position is never achieved during the motion. In fact, both the torque policy and PD policy are used for torque control on the real robot.

\subsubsection{Structured Action Spaces}
Instead of directly controlling the robot via joint space control, many works also explore controlling the robot in the task space (feet for instance),  e.g.,~\cite{krishna2022linear, duan2021learning, castillo2023template}. This allows for boosting learning efficiency and simpler control architecture. However, direct control in joint space still dominates the literature, mainly because it is more general and avoids singularities when inverse problems are solved. 

Prior knowledge of how the robot should move can also be embedded in the action space via the use of residual RL~\cite{johannink2019residual}. In this approach, an open-loop reference control signal $\hat{\mathbf{a}}_t$ is provided, and policies are learned to generate feedback signals, which are then added to the reference. For example, in \cite{iscen2018policies}, reference signals $\hat{\mathbf{a}}_t$ is a sinusoidal wave over time that encourages the policy to generate desired gaits such as trotting and bounding, and the sum $\mathbf{a}_t = \hat{\mathbf{a}}_t + \pi(\mathbf{s}_t)$ are used as the joint target for the low-level PD policy.

One can also learn to modulate parameters of high-level motion primitives. For example, ~\cite{bellegarda2022cpg} learn to modulate the intrinsic oscillator amplitude and phases of a central pattern generator. ~\cite{xie2022glide, margolis2021learning} learns to output the desired center of mass accelerations, which are then used to generate motor commands using the single rigid body model. We will further detail these approaches in Sec. \ref{sec:combined}.

\section{Learning Frameworks}\label{sec:frameworks}
In the previous section, we discussed the common formulation of MDPs for legged locomotion. Solving the MDP is a highly challenging non-linear optimization problem and thus there has been extensive research to efficiently solve this problem. In this section, we discuss popular choices of learning frameworks researchers have explored to solve the MDP for locomotion policies.

% Theoretically, the formulated MDP can be solved with any off-the-shelf MDP problems in an end-to-end fashion. However, learning often suffers from many practical issues, including sample inefficiency, poor convergence, or bad sim-to-real transfer, which provides motivations for specialized learning frameworks. This section aims to provide an overview of popular choices, such as curriculum learning, hierarchical learning, and privileged learning, which have been adopted by numerous legged locomotion researchers. Note that these techniques are not mutually exclusive and can be curated in many different ways.  

\subsection{End-to-end Learning}
The most straightforward approach would be to treat the given MDP as a monolithic formulation and solve everything end-to-end using DRL algorithms. Among diverse choices, the most popular DRL algorithm for solving legged locomotion is an on-policy algorithm, such as TRPO \cite{schulman2015trust} and PPO \cite{schulman2017proximal}. These algorithms are designed to take conservative update steps by constraining the parameter changes within the \emph{trust region} defined in the vicinity of the current policy. As a result, these on-policy algorithms provide a robust learning framework that can reliably find high-performing policies at the end, which makes them suitable for challenging locomotion problems. However, off-policy algorithms, such as DDPG \cite{lillicrap2015continuous} or SAC \cite{haarnoja2018soft}, have also been adopted by many researchers, particularly when the sample efficiency becomes a bottleneck of the problem.

End-to-end learning works well when the initial policy can effectively explore states that provide learning signals. However, for more challenging tasks and without any structure in the action space, the initial policy may not be able to obtain any useful reward signal and thus make it difficult for end-to-end methods to work well. In such cases, adding more problem-specific structures to the problem can be helpful, e.g., curriculum and hierarchical learning.

\begin{figure}
    \centering
    \includegraphics[trim={0cm 1.0cm 0cm 1.0cm},clip,width=1.0\linewidth]{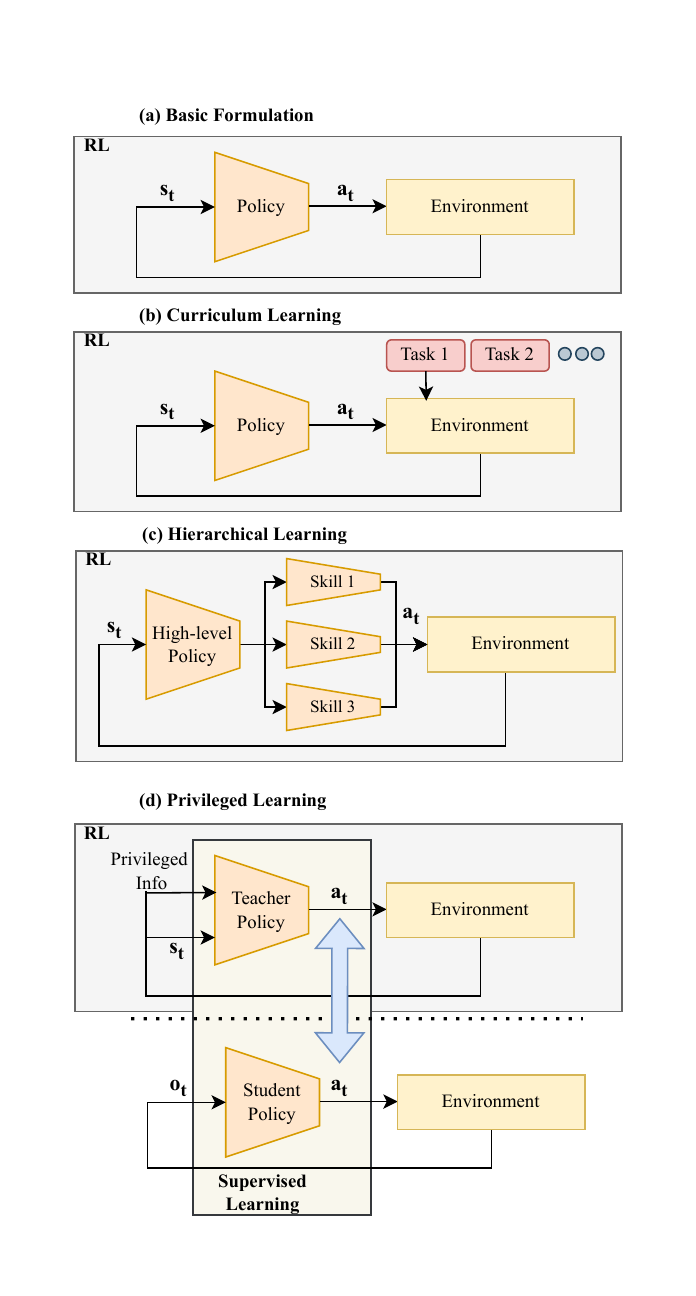}
    
    \caption{Illustration of popular learning frameworks: (a) basic learning, (b) curriculum learning, (c) hierarchical learning, and (d) privileged learning.
    }
    \label{fig:frameworks}
\end{figure}

\subsection{Curriculum Learning}
Similar to the way schools create progressively challenging learning milestones to enhance educational outcomes, researchers have also explored a similar idea known as curriculum Learning (CL) for handling more difficult problems in robot learning.
CL can be applied to different aspects of the learning framework. For example, \cite{heess2017emergence, xie2020allsteps, rudin2022learning,margolis2024rapid} trained virtual and real agents in increasingly challenging environments and obtained policies that can handle environments that basic learning cannot achieve. CL has also been applied to improve the robustness of the policies by imposing an increasing amount of disturbances and randomness throughout the training \cite{akkaya2019solving}. Constraints can also be enforced in a CL setting by starting with a soft constraint and gradually tightening it \cite{zhuang2023robot}.

An important aspect of CL settings is to decide when to progress to the next stage of training and what the next stage should be. For locomotion problems on uneven terrains, the parameters used for terrain generation can parameterize the stages of the curriculum, e.g., the steepness of a slope or the height of a stair~\cite{rudin2022learning}. The stages of a curriculum can then be manually defined and the decision to progress to the next stage is usually determined by the capability of the policy, as measured by the sum of rewards it gets over one episode~\cite{xie2020allsteps} or the progress it can make over the current stage~\cite{rudin2022learning}. Instead of manually defining the next stage of the curriculum, adaptive curriculum methods are also proposed to update the sampling distribution of the curriculum parameter. For example in ~\cite{xie2020allsteps}, the authors propose an adaptive curriculum strategy to update the sampling distribution by estimating the capability of the policies over a range of terrain parameters as measured by the value function. To enable high-speed locomotion skills for a mini-cheetah robot, \cite{margolis2024rapid} propose to adapt the sampling distribution of the velocity commands using the rewards it gets.

\subsection{Hierarchical Learning}
Another common strategy to handle more challenging learning problems, especially ones with long horizons like navigation, is to decompose the problem into hierarchies. Typically, the problem is decomposed into high-level tasks and low-level skills, where the action space of the high-level task will be part of the input of the low-level skills, and each level is learned separately.  

Hierarchical learning is adopted in several learning-based control works in the computer graphics literature. The decomposition of the problem can be based on human intuition, for example, in \cite{peng2017deeploco}, a high-level policy will prescribe the desired footsteps to accomplish tasks such as navigation and soccer dribbling, while the low-level policy takes the desired footsteps as input to generate joint level control to follow the footsteps. The output of the high-level policy can also be a learned latent space. For example, \cite{peng2022ase, won2022physics, zhu2023neural} present frameworks where the low-level policies are trained to reproduce diverse motion capture data, driven by a learned latent representation; high-level policies are then trained to manipulate these latent spaces to accomplish high-level tasks.

Similar frameworks are adopted in several quadrupedal locomotion works. \cite{han2023lifelike} learn low-level policies using the vector quantized variational encoder (VQ-VAE) \cite{van2017neural}. This allows a quadruped to reproduce a variety of motion capture data of animals and a high-level policy to manipulate the latent space of the VQ-VAE to realize navigation and multi-agent chase tag game. \cite{ji2022hierarchical} divide a soccer shooting task into an end-effector motion tracking task and an end-effector trajectory planning task, where the output of the trajectory planning is used as input for the motion tracking policy. The two tasks are learned separately to accomplish quadruped soccer shooting.  \cite{ji2023dribblebot} similarly learn task and recovery policies for soccer dribbling.

There are also hierarchical architectures combining a model-based control approach and a learning-based approach, where a control policy can either be learned or manually designed. We leave the discussion to Sec. \ref{sec:combined}. 

\subsection{Privileged Learning}\label{subsec:privileged_learning}

Robotic tasks in the real world are inherently partially observable, leading to challenges in achieving high-performance locomotion. For instance, accurately measuring or estimating factors like payload weight or friction coefficient is nearly impossible, but crucial for the policy to be robust or adaptive.

To address these challenges, existing approaches often involve constructing a belief state from a history of available measurements. In DRL, this is typically accomplished by stacking a sequence of previous observations~\cite{haarnoja2018soft, hwangbo2019learning} or using models with memory such as RNN or TCN.
However, training a complex neural network policy with such a large input space using RL from scratch can be time-consuming and challenging.

\cite{chen2020learning} propose a strategy called \emph{learning by cheating}, which was renamed to \emph{privileged learning} in the context of quadrupedal locomotion \cite{lee2020learning}. This approach leverages fully observable MDP in simulation to train a \emph{teacher} policy, which is then distilled into a \emph{student} policy based on sequence models like RNN or TCN. 
Specifically, the teacher has access to privileged information, such as the noiseless, rich simulated states of the environment (e.g., friction coefficient). In this context, \cite{chen2020learning} collected human demonstrations in simulation, while \cite{lee2020learning} trained a teacher policy for quadrupedal locomotion using noiseless robot states and terrain information as privileged information. At test time, the student policy has no access to such information, instead, it processes a history of measurements available from the robot. 

\cite{lee2020learning} first demonstrated the effectiveness of this approach for quadrupedal locomotion, enabling rough terrain traversal that outperformed conventional optimization-based approaches.
By imitating the privileged teacher policy, the student policy constructs an internal representation of the world that conveys locomotion-related information.
This idea of combining RL and imitation learning has been adopted by subsequent works to enhance robustness or utilize more complex input modalities such as raw images or voxel maps~\cite{miki2022learning,haarnoja2024learning,agarwal2023legged,miki2024learning,lee2024AOW}.

\subsection{Starting Point to Train Your Robot}
Up to this point, we have provided all the information required to train a legged robot in simulation. To further help those new to the field, we give some further recommendations on where to start. In terms of algorithms, there are simple and intuitive repositories that explain (mainly RL) algorithms with code, e.g., \cite{abel2019simple_rl,huang2022cleanrl}. In terms of simulation, as suggested by Fig. \ref{fig:sim}, Isaac and Mujoco are the best choices to start with. In particular, Mujoco open-sourced the simulation code which can help with understanding the components within a simulator. A wide range of URDFs for robots can be found in \cite{caron2022awesome}.\newline
Starting from a robot model, one can build an environment for reinforcement learning. An initial set of observations can include base height and orientation, base linear and angular velocity, and joint angles and velocities. The action space can be set to be the target joint angle for a joint PD controller and the reward can be designed around tracking a desired body velocity. PPO~\cite{schulman2017proximal} is now the most popular algorithm for training locomotion policies for legged robots and there are many open-source implementations, e.g., \cite{rudin2021rsl, sidor2018stable}.
Several open-source codebases for learning quadruped locomotion are available. Some representative choices are listed here for the readers' reference: the implementation of \cite{peng2020learning} can be found in \cite{coumans2020motion}in the Pybullet simulator \cite{coumans2016pybullet}. It also contains an MPC-based controller similar to \cite{bledt2018cheetah}. \cite{rudin2021isaac} contains the implementation of~\cite{rudin2022learning} in Isaac Sim~\cite{makoviychuk2021isaac}. Most recent works on legged robot learning are built on top of this, thanks to the efficiency of Isaac Sim.

\section{Sim-to-Real Transfer}

The promise of computer simulations to efficiently generate large amounts of training data creates a great synergy with modern deep learning algorithms that are capable of absorbing large-scale datasets and often require abundant data sources to achieve effective learning. However, a major challenge for policies that are trained using simulation data is whether they will perform well on real hardware due to the discrepancies between computer simulation and real-world robot dynamics, also known as the sim-to-real gap. It is not practical to construct a simulation model to accurately represent every aspect of the real world due to its complexity. For example, motors can heat up over time, showing a different behavior given the same command input, non-deterministic delays exist due to communication of different parts of the hardware, and the feet of the robots are usually made of soft materials instead of the rigid body model commonly used in most simulators. Furthermore, a policy trained entirely in simulation can take advantage of some aspects of the simulation that are not present in reality, resulting in direct failure upon direct deployment. In this section, we review some of the techniques commonly used to combat the sim-to-real gap.

\subsection{Good System Design}
A naive implementation of the environment for locomotion tasks often leads to undesirable behaviors that take advantage of the peculiarities of the simulators. For example, a reward that encourages forward walking motion for the commonly used MuJoCo humanoid often generates control policies that resemble bang-bang control, with the joints oscillating at high frequency, resulting in behaviors that glide forward. Such a bang-bang control strategy is not suitable for hardware and sim-to-real is doomed to fail. A good design of the overall system helps constrain the policies to operate in the same distribution as the simulator and to combat the sim-to-real gap.

\paragraph{Reward Design} The design of the reward function can directly affect the sim-to-real performance. In Sec. \ref{subsec:reward}, we described the most common reward terms, but we reiterate some important elements here. To avoid jittery motions, joint acceleration are often penalized; a reward for foot air time is often used to avoid foot-dragging behavior; foot impact penalty is used to avoid stomping behaviors \cite{makoviychuk2021isaac}. Maintaining a good balance between task rewards such as velocity tracking and regularization rewards makes the reward shaping problem a laborious task.

\paragraph{Observation and Action Space Design} The choice of observation and action space also plays an important role. \cite{xie2021dynamics} identify several key factors that enable direct sim-to-real transfer for the Laikago robot, without randomization or adaptation techniques commonly used in other works. In particular, a low proportional gain for the joint PD controller that allows for compliant behavior greatly reduces sim-to-real gap, and the use of a state estimator that allows for using body velocity as part of the observation allows for rejecting velocity perturbation.

\paragraph{Domain Knowledge} Domain knowledge can also be used to improve the system design. For example, in \cite{xie2020learning}, symmetry constraints are used to promote the left-right symmetry of bipedal walking policies, greatly improving the motion quality and as a result greatly improving sim-to-real performance. Domain knowledge can also be used in the reward function to promote certain behaviors, for example, the use of motion capture data in the reward function can promote the style of the motions \cite{peng2020learning, han2023lifelike} or using CPGs as a structure in the policy \cite{lee2020learning, bellegarda2022cpg, iscen2018policies} can lead to more desirable gait patterns.

With good system design, a few works demonstrated sim-to-real success for legged robot locomotion without dynamics randomization. For instance, \cite{smith2023learning} demonstrate successful sim-to-real transfer of highly dynamic motions (such as jumping and walking on hind legs) on the A1 quadruped. \cite{xie2020learning, dao2020learning} successfully show sim-to-real for the bipedal robot Cassie. These examples demonstrate that a good design can sometimes help bridge the sim-to-real gap, without a need for any further strategies.

% \begin{table}[]
%     \centering
%     \begin{tabular}{c|c}
%         \hline
%         design & effect \\ \hline
%         reward tuning &  good motion quality  \\ \hline
%         domain knowledge & good motion quality \\ \hline
%         soft proportional gain & compliant behaviors \\ \hline
%          observation & perturbation rejection                                          \\
%     \end{tabular}
%     \caption{Some designs that can improve sim-to-real transfer.}
%     \label{tab:design}
% \end{table}

% For example, \citet{xie2021dynamics} presents a system to directly train control policies in simulation using RL for the Laikago robot and transfer them successfully to the real hardware without additional sim-to-real techniques like dynamics randomization or adaptation. They identify several design decisions that are crucial to the sim-to-real success, including the proportional gain used for the joint PD controller and the inclusion of body velocity in the observation space. Many other works also demonstrate successful sim-to-real with well-designed systems, e.g., \cite{xie2020learning}, where an iterative design process is introduced to refine control policies, resulting in successful sim-to-real transfer for the Cassie biped, \cite{smith2023learning} demonstrates successful sim-to-real transfer for the A1 quadruped for highly dynamics motion such as jumping and walking on hind legs. \sehoon{The takeways from this subsection are a bit vague. We may want to enumerate design suggestions explicitly.}

\subsection{System Identification}
Even with good system design, inaccurate modeling or unmodelled dynamics in the physical system can cause direct sim-to-real to fail. To improve the fidelity of the simulator, system identification can be performed.
A dominant source of modeling errors comes from the actuator dynamics. In simulation, motors can apply arbitrary torque profiles that a policy commands, while a physical actuator often produces less accurate torque profiles due to the limited bandwidth as well as limited tracking accuracy of the underlying motor controller. One of the earliest successful sim-to-real transfer for quadrupedal locomotion used an analytic actuator model~\cite{tan2018sim} for a miniature quadruped. Later~\cite{hwangbo2019learning} used a fully black-box model (neural network) to learn an actuator model, which was then used in the simulation to train locomotion policies. 

Another important source of the sim-to-real gap is the discrepancy between the contact model in simulation and the real-world interactions. Most of the existing simulators use rigid contact, while in reality there always exist some deformations in the interaction. However, it is quite challenging to simulate deformable terrains or those with complicated shapes, which often involves prohibitively expensive finite-element methods. One notable success in using a more complicated contact model is \cite{choi2023learning}, which develops a compliant contact model to simulate diverse terrains, ranging from very soft beach sand to hard asphalt. Training with this model boosts the performance of the policy handling similar terrains in the real world compared to training with a rigid contact model. 

\subsection{Domain Randomization}

In addition to better system design and more accurate simulation, another important strategy to mitigate sim-to-real gap is to improve the generalization capabilities of the trained policies. One important approach to improve policy transfer is domain randomization, where the system parameters are randomized during training. This is similar to robust control, where a controller is optimized to perform robustly under a bounded set of system parameters and uncertainties \cite{dorato1987historical}. A fundamental assumption behind domain randomization is that if we generate control policies that can handle sufficiently diverse training environments, it is more likely that this policy also work in the real world, even though it has never seen it during training. Another way to interpret this approach is that if we have a control policy that works for an ensemble of different environments, the real world will hopefully fall into one of these environments, hence the policy will work in the real world.

Early works in using domain randomization for robot learning applied the idea to the problem of manipulation to enable more robust perception modules \cite{tobin2017domain}. In the legged locomotion field, \cite{tan2018sim} were amongst the first to adopt the idea of domain randomization. Specifically, they randomized key dynamic parameters that are important for obtaining robust policies such as robot mass, friction coefficient, motor strength, and latency. This approach has since been widely adopted in follow-up works and combined with feed-forward trajectories \cite{iscen2018policies, lee2020learning, miki2022learning}, motion imitation \cite{peng2020learning}, task-space control \cite{bellegarda2022robust}, and policy distillation \cite{caluwaerts2023barkour, kumar2021rma} to further improve legged locomotion learning in different scenarios.

The aforementioned approach in domain randomization largely focused on bridging the physics gap between the simulation and the real world, where the resulting policies are often blind to their environments. As these techniques mature, recent works in legged locomotion explored randomizing also the simulated visual perception parameters such as camera intrinsics and extrinsics, and noise models to achieve reliable vision-based legged locomotion on highly unstructured terrains \cite{yu2021visual, miki2022learning, margolis2023walk, zhuang2023robot, agarwal2023legged, cheng2023extreme}.
Notably, \cite{truong2023rethinking} reported that higher visual fidelity does not always lead to better performance due to slow simulation speed.

\subsection{Domain Adaptation}
\begin{figure}
    \centering
    \includegraphics[width=\linewidth]{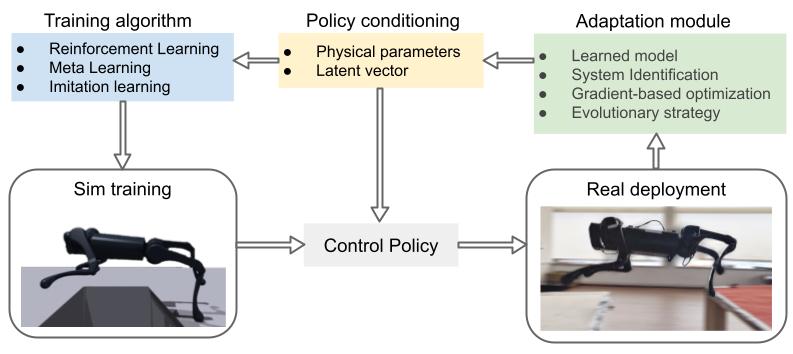}\
    \caption{Common components in domain adaptation algorithms \cite{zhuang2023robot}.}
    \label{fig:domain_adaptation}
\end{figure}

Like domain randomization, domain adaptation aims to develop generalizable policies that can cover real-world environments, with the main difference being that domain adaptation explicitly identifies the current scenario that the policy is operating within and adjusts the policy behavior to be optimal for the current scenario (similar to adaptive control). As a result, when successfully trained, domain adaptation-based policies can often achieve better performance and cover a wider range of scenarios than domain randomization techniques. 
Figure~\ref{fig:domain_adaptation} demonstrates common components in domain adaptation algorithms. 

One class of domain adaptation algorithms is to explicitly identify the parameters of the environment, which are used as input to the control policy or during policy training \cite{Yu-RSS-17, chebotar2019closing, yu2019sim, muratore2021data, lee2018bayesian}. To identify the environment parameters, \cite{Yu-RSS-17} learned models,  \cite{chebotar2019closing} optimized trajectory matching loss, and \cite{yu2019sim} directly optimized task performance. Although explicitly identifying the environmental parameters enables an intuitive interpretation of the adaptation process, it is difficult to extend this idea to cases where a moderate number of environment parameters need to be adapted.

To mitigate this issue, researchers have opted for implicitly representing the environment parameters during the adaptation \cite{yu2020learning, kumar2021rma, peng2020learning, lee2020learning}, where the high-dimensional environment parameters are compressed into a low-dimensional latent representation. There is a variety of approaches for obtaining a good latent representation. For example, \cite{peng2020learning} propose an optimization-based approach to find the latent environment representation for an estimated advantage value during policy adaptation. \cite{kumar2021rma} and \cite{lee2020learning} on the other hand, directly learn the latent environment representation during policy training, and later on train a separate system identification module to predict the latent representation during inference. \cite{lee2022pi} leverage ideas from the representation learning field and use predictive information to acquire a state representation that captures key environment dynamic information. Due to the ability to handle larger amounts of environment variations, this class of domain adaptation approaches has been more widely applied to real legged systems and has become one of the most popular classes of methods in recent years.

There are also works that learn to improve the simulation accuracy by using real-world data. The transition function in a simulation can be augmented in order to better match the real world. Policies trained with the augmented simulation are shown to demonstrate better sim-to-real performance, e.g., \cite{golemo2018sim,jeong2019modelling,jiang2021simgan}.

\section{Combining Control and Learning}\label{sec:combined}
While both RL and OC aim at finding the optimal policy for a desired task, they have complementary features that attract researchers to devise frameworks that benefit from the best of both worlds. For instance, as model-based approaches use the knowledge of dynamics and constraints in the policy design, they transfer our knowledge of the world among different tasks. Furthermore, they can explicitly take constraints into account which makes them suitable for safety-critical cases. On the other hand, learning-based approaches can randomize the parts of the model with uncertainties and deliver a robust policy without much overhead in the policy design. Furthermore, the online computation is at a minimum, as training is done purely in simulation and one only needs to perform inference during execution.

This line of work can be split into four general categories; 1) learning some parameters of model-based controllers, 2) learning high-level policies to set commands for a model-based controller, 3) leveraging learning to improve the generalization and reduce online computation of model-based controllers, 4) using model-based control to make learning more sample-efficient.

\subsection{Learning Control Parameters} 
There are certain parameters in model-based control policies that directly affect the performance and robustness of the controller. Normally, tuning these parameters is done manually on the real system by the control designer. However, it is always desired to automate this process. To do that, \cite{ponton2014learning,heijmink2017learning} use PI$^2$ to learn optimal gains for tracking a desired trajectory. \cite{pandala2022robust} learns the boundaries of the uncertainty set that is then used to design a robust MPC controller.

To efficiently learn the control parameters, some recent approaches rely on Bayesian optimization (BO). 
% Early works within this category focused on learning the cost of an instantaneous whole-body controller. 
% \cite{charbonneau2018learning} learns the task priorities of a QP-based whole-body controller.  \cite{yeganegi2019robust,yeganegi2021robust,shahrokhshahi2022sample} learn the cost function of a linear MPC to generate plans that are then tracked using a whole-body controller. 
\cite{yeganegi2019robust,yeganegi2021robust} learn the cost weights of the trajectory optimization module and an MPC controller for humanoid locomotion, respectively.
\cite{marco2021robot} learn the joint impedance to track trajectories from a trajectory optimization module, while simultaneously learning the failure constraints. \cite{yang2022bayesian} use BO to learn the control parameters of a hybrid-zero-dynamics-based controller. As BO can only handle a small number of decision variables, \cite{yuan2019bayesian,sarmadi2023high} first perform dimensionality reduction and then apply BO to the low-dimensional latent space. To make the exploration in the real world safe, \cite{widmer2023tuning} apply a safe BO framework to tune both the gait and feedback control parameters in the real world.

\subsection{Learning a High-level policy}
A more sophisticated version of simply learning a few parameters of a model-based controller is to learn a policy that works in concert with a model-based controller. For instance, given a general map of the environment, finding foothold locations for locomotion is a highly challenging task for model-based controllers. Furthermore, model-based controllers are sensitive to the precision of their underlying model, hence a residual policy can correct for this model discrepancy.

\cite{villarreal2020mpc,byun2021learning,xu2022learning,omar2023safesteps} learn a contact planner that, given the terrain map, finds safe foothold locations and passes them to an MPC module to optimize ground reaction forces. \cite{gangapurwala2022rloc} train a policy that generates desired foot locations which are tracked using a model-based whole-body controller. \cite{taouil2023quadrupedal} develop a footstep location planner for a given low-level controller that is conditioned on a desired velocity. \cite{dhedin2024diffusion} leverage the power of diffusion models in handling multi-modal datasets to learn in a supervised fashion a policy that outputs contact plans for MPC. 
\cite{gangapurwala2021real} leverage DRL to learn a feedback trajectory corrector to update the output of trajectory optimization that is then stabilized using a whole-body controller. \cite{yang2022fast} learn a high-level policy to transition between different gaits.  \cite{viereck2021learning,xie2022glide} learn a centroidal trajectory planner that generates desired forces and body trajectories that are fed to an inverse dynamics controller for tracking. 

\subsection{Learning for Efficient Model-based Control}
While considering the model in the structure of the policy can result in a better generalization, it comes with the cost of more online computational burden. For instance, a model-based controller needs to plan online for a walking motion in real-time. However, it seems much more efficient to simply cache the solutions in an offline phase, e.g., using a function approximator, to reduce the online computation. In this vein, \cite{carius2020mpc,reske2021imitation} learn the control Hamiltonian function, whereas  \cite{wang2022learning,viereck2022valuenetqp,kovalev2023combining} learn the value function of the optimal control problem. \cite{lembono2020learning,dantec2021whole} use a large dataset from trajectory optimization to learn warm-starts for quick resolutions of the nonlinear program in the MPC. \cite{yang2020data,bechtle2021leveraging,anne2021meta,li2021planning,wu2023daydreamer} learn a forward model that is then used within a model-based module to control the robot. \cite{kwon2020fast} learn the solution map of the trajectory optimization problem to enable reactive planning in real-time.

\subsection{Model-based Control for Efficient Learning}
The fourth category leverages tools from model-based control to efficiently learn locomotion skills. For instance, one can use trajectory optimization to guide the RL through high-reward parts of state space, for environments with sparse rewards. The final policy in these approaches is a neural network that sends actions to the joins.

Within this line of research,
\cite{gangapurwala2020guided} use optimal control to guide and constrain the exploration of the DRL algorithm. \cite{tsounis2020deepgait} leverage trajectory optimization to evaluate gait transitions in the high-level controller in a hierarchical DRL setting. \cite{green2021learning} leverage the power of reduced-order models for planning and learn a residual policy to compensate for the discrepancy between the reduced and full robot models. \cite{bellegarda2020robust,bogdanovic2022model,brakel2022learning,kang2023rl+} use optimal control to generate demonstrations to guide DRL. Similarly, \cite{jenelten2024dtc} interactively queries a trajectory optimization module to provide demonstrations for the DRL policy in different initial and final configurations. \cite{fuchioka2023opt} investigate the use of feed-forward inputs from trajectory optimization to improve the RL learning efficiency and sim-to-real transfer. \cite{khadiv2023learning} take a behavioral cloning approach to learn optimal policy from MPC directly in sensor space, while \cite{pua2024safe} introduce a more sample-efficient framework compared to behavioral cloning. \cite{youm2023imitating} use a similar approach, but improves the cloned policy using another DRL phase like \cite{bogdanovic2022model}. 

\section{From Quadrupeds to Bipeds}
Works using RL for bipedal locomotion dates back to as early as 2005, when policy gradient \cite{tedrake2005learning} was used to train a small biped to walk in under 20 minutes. \citet{schuitema2010design} use RL to train a 2D biped to walk in 2010. \citet{hester2010generalized} applied model-based RL to learn soccer penalty kicks for a NAO humanoid. These early works focused on learning directly on the physical robots and therefore were only applied to simplified bipeds, e.g., a small biped with large feet or a biped with motion restricted to a 2D plane.

From early 2000 to the \cite{DRC2015final} (DRC) in 2015, the dominant approach to control humanoid robots was OC or heuristic methods with simplified dynamics models. Early works on the use of OC for humanoids resorted to the simplified linear models of the dynamics to enable linear MPC \cite{kajita2003biped,wieber2006trajectory}. However, for multi-contact behaviors (making contact with the environment using both feet and arm), later works focused on solving the holistic problem using the robot full dynamics with contact \cite{tassa2012synthesis,mordatch2012discovery,lengagne2013generation,posa2014direct}. These holistic approaches showed little success through a model predictive control fashion in real-world humanoid control \cite{koenemann2015whole}. In DRC 2015, most teams used a combination of simplified models planning together with full-body inverse dynamics/kinematics~\cite{feng2015optimization,kuindersma2016optimization,johnson2017team}. This DRC demonstrated that humanoid technology is by no measure close to being deployed for real-world problems and a paradigm shift is required~\cite{murphy2015meta,atkeson2018happened}.

As DRL achieved success in physics-based character animation~\cite{peng2018deepmimic} and quadrupedal locomotion~\cite{tan2018sim,hwangbo2019learning}, researchers started to use DRL algorithms in physics simulations to train walking policies for the bipedal robot Cassie and successfully transferred them to the hardware~\cite{xie2020learning}. Techniques such as domain randomization, more modern neural network architectures, and smarter reward design enabled dynamic and versatile behaviors that are robust to real-world perturbation, e.g., \cite{siekmann2021sim, radosavovic2024real,li2024reinforcement}. More recent works explored how to incorporate perception to enable challenging terrain navigation, such as stepping stones~\cite{duan2022learning} and uneven terrains~\cite{duan2023learning}.
In addition, more expressive learning-based motions are recently shown on humanoid robots~\cite{cheng2024expressive}.

In the years 2023 and 2024, we witness an explosion of new companies starting to develop humanoid robots, most of which have bipedal forms. The highlight of the exhibition section of the 2024 International Conference in Robotics and Automation (ICRA) was an immense increase in the number of electrically-actuated humanoids. In particular, Unitree has unveiled its new humanoid G1 with an incredibly low price, on par with manipulator arms. This marks the starting point of a new era where many research laboratories will have access to humanoid robots. We have seen this trend in the past decade that enabled exponential progress in the fields of drones and quadrupedal locomotion. Furthermore, due to the simplicity and ease of use of modern DRL algorithms and the availability of fast and parallelizable simulators, the barriers to generating and implementing new motions on highly complex humanoid robots are being removed. This has resulted in a new wave of pioneering demos from the industry, e.g., backflip motion from Unitree H1 \cite{unitree2024h1} and highly stylish motions of the Disney bipedal creature \cite{disney2023biped}. We believe this trend will continue and we will see more and more of these impressive behaviors on real robots from different industrial players.

Parallel to the humanoid efforts in industry, researchers have developed open-source \cite{daneshmand2021variable} and affordable \cite{liu2022design} miniature bipedal platforms. These miniature robots are much easier to handle than full humanoid robots and could potentially impact the rate of progress of research on learning-based bipedal locomotion. For instance, \cite{haarnoja2024learning} develop a hierarchical system to enable learning of soccer plays on the miniature Robotis OP3 humanoid. Recently, the miniature robot from LimX has shown impressively robust bipedal locomotion behaviors in the wild \cite{LimX2024biped}. With affordable human-size humanoids becoming widely available, one would expect lessons learned from quadrupedal learning works to be transferred to humanoid robots, while also additional techniques are to be developed to handle the difficulties of balancing bipeds compared to quadrupeds.

\section{Unsolved Problems and Research Frontiers}
%  Ongoing Research and Future Directions
As surveyed in the previous sections, there have been major developments in the field since the early work on the use of DRL for legged robots \cite{tan2018sim,hwangbo2019learning}.  The significant progress enabled by learning-based approaches opens the door to new challenges. One major issue of current DRL algorithms is their need for heavy reward shaping for each skill and the lack of transfer of knowledge between skills. Unsupervised skill discovery is an interesting avenue for more research in this regard (Sec. \ref{sec:unsupervised}). Another issue is the sample inefficiency of DRL algorithms, which begs for the use of higher-order gradients in policy optimization. Section \ref{sec:diff_sim} discusses how differentiable simulators can be leveraged to improve sample efficiency.\newline
While DRL has enabled many robust behaviors in the wild, they struggle in safety-critical situations such as stepping stones and confined environments (Sec. \ref{sec:challenging_env}). This will require handling safety constraints which is an interesting area for future research (Sec. \ref{sec:safety}). Furthermore, the deployment of legged robots in the real world may need more efficient mobility. Exploring hybrid locomotion modalities is another interesting aspect that needs more research (Sec. \ref{sec:wheel_leg}).\newline
As legs are becoming the modality of choice for traversing various environments, locomotion needs to be combined with manipulation to perform tasks in the real world. Adding the complexity of object manipulation to the locomotion problem opens up new challenges that require a substantial amount of research (Sec. \ref{sec:loco-manipulation}). Finally, the rise of foundation models has sparked interest among roboticists. In particular, the fact that these models have shown some capability for common-sense reasoning, points to them as being potentially useful tools to handle the combinatorial nature of the loco-manipulation problem in the real world (Sec. \ref{sec:foundation_models}).

\subsection{Unsupervised Skill Discovery}\label{sec:unsupervised}
DRL often requires labor-intensive reward engineering of domain experts to obtain desirable behaviors. Instead, unsupervised skill discovery can potentially reduce the burden of repetitive reward engineering by learning a set of reusable skills based on intrinsic motivation. Typically, unsupervised skill discovery aims to learn a skill-conditioned policy that executes diverse and discernable skills, as discussed in \cite{eysenbach2018diversity}.

Researchers have applied the idea of unsupervised skill discovery to learn diverse locomotion skills~\cite{sharma2020emergent,cheng2023learning}. For example, \cite{schwarke2023curiosity} successfully used this approach to learn loco-manipulation skills for a real quadrupedal robot. Once the skills are obtained, they can be leveraged for downstream tasks via model-predictive control~\cite{sharma2020emergent} or hierarchical RL~\cite{eysenbach2018diversity}. A similar idea has been also applied to learn skills from unstructured motions from expert policies~\cite{li2023versatile} or even from human motion clips~\cite{li2023crossloco}. 

\subsection{Differentiable Simulators}\label{sec:diff_sim}
To improve the sample efficiency of learning-based approaches for legged locomotion, there has been an increasing interest in the use of differentiable simulators \cite{schwarke2024learning}. The main idea in these approaches is to better exploit the structure available in the simulation environments, for instance through the use of gradients obtained either analytically \cite{werling2021fast,howell2022dojo} or using automatic differentiation \cite{degrave2019differentiable,hu2019difftaichi,freeman2021brax} or even finite difference \cite{todorov2012mujoco,mujoco2023jax}. However, the main problem with this approach is poor local minima due to contact. One interesting approach to overcome this issue is the use of randomized smoothing techniques \cite{suh2022bundled,lidec2022augmenting,pang2023global}.  As more and more differentiable simulators become available, this under-explored area in the field can attract more researchers.

\subsection{Traversing Challenging Environments}\label{sec:challenging_env}
After the success of DRL in generating highly robust behaviors \cite{hwangbo2019learning,lee2020learning,bogdanovic2022model}, recently there has been an increasing interest in learning locomotion skills in challenging environments, e.g., performing parkour \cite{zhuang2023robot,cheng2023extreme,hoeller2023anymal}, traversing stepping stones
\cite{duan2022learning,zhang2023learning,dhedin2024diffusion}, moving in confined environments
\cite{xu2024dexterous,chane2024cat} (see Fig. \ref{fig:snapshots} for a few examples). This is an interesting direction that can greatly benefit from the power of deep learning to include perception in the control loop. While it has been argued that DRL-based controllers may struggle with traversing highly constrained environments \cite{grandia2023perceptive}, recent efforts have shown the opposite \cite{zhang2023learning}. However, it remains an interesting question to be answered; to what extent DRL approaches can handle safety-critical situations?

\subsection{Safety}\label{sec:safety}

With continuous advancements in robot locomotion, they will start to be deployed in more diverse locations such as homes, offices, plants, or in the wild. As such, safety becomes a critical topic as we want to make sure the robots do not damage themselves during deployment nor do they damage the environment or any human being nearby. One possible way to tackle safety in robot locomotion is to formulate it as a constrained optimization problem. Indeed, researchers in safe reinforcement learning have developed a variety of safe RL algorithms that aim to train control policies without violating safety constraints \cite{thananjeyan2021recovery, hsu2023safety, achiam2017constrained, liu2022constrained}, where they focus on theoretical properties of the algorithms and are applied to synthetic tasks. More recently, researchers have started to apply these methods to legged robots and demonstrated safe learning in the real-world \cite{yang2022safe}. Constrained policy optimization approaches have also been used to achieve safer and more precise behaviors for locomotion in complex environments \cite{kim2023not, lee2023evaluation, xu2024dexterous}. It is anticipated that this direction of research will continue to thrive with the more advanced legged robot capabilities and hardware. 

Another way to ensure safety, specifically in deployment, is to use safety filters, which have been extensively studied in the control community \cite{wabersich2023data}. These filters can be designed using  Hamilton-jacobi (HJ) reachability, control barrier functions (CBF), and predictive methods for uncertain systems. Variations of algorithms with and without access to the robot dynamics have been developed. However, their application to locomotion has been limited \cite{kim2023safety,he2024agile}. More research on this under-explored area in the future can facilitate certification of this technology for the real-world safety-critical applications.

\subsection{Hybrid Wheeled-legged Locomotion}\label{sec:wheel_leg}
A wheeled-legged robot is a variation of legged robots designed to enhance efficiency and endurance.
Unlike traditional point-foot legged robots, the actuated wheels allow for an additional mode of locomotion, enabling them to drive without stepping in the longitudinal direction. 

Achieving both efficiency and stability requires a locomotion controller capable of transitioning between walking and driving locomotion modes.
In conventional model-based approaches, this mode switching often relies on operator commands or heuristics \cite{bjelonic2019keep, bjelonic2020whole}. However, gait generation for wheeled-legged robots is not straightforward, because there is no such systems in the nature.  This makes heuristic approaches with periodic patterns such as CPG impractical.

While this research field is still very young, recent studies have explored training locomotion policies in an end-to-end manner using DRL.
\cite{lee2024AOW} applied an approach similar to \cite{miki2022learning} without CPGs to a wheeled-legged system, and demonstrated adaptive gait transitions and locomotion over rough terrains. Similarly, \cite{chamorro2024reinforcement,cui2021learning} demonstrated wheeled-legged bipeds traversing uneven terrains, albeit without gait transitions. This combined modality has a lot to offer and it seems to be an interesting direction for more research.

\subsection{Loco-manipulation}\label{sec:loco-manipulation}
When manipulating objects by a legged robot, it becomes essential to take the object affordances, dynamics, and constraints into account in the planning and control phases. Given the affordances of the objects and the environment \cite{gibson1977theory}, the robot needs to simultaneously decide on both locomotion and manipulation aspects. There have been recent efforts on the use of DRL for loco-manipulation tasks in the real world. In general, there are two ways to think about loco-manipulation using quadrupeds. The first one is to use the robot's body and feet to perform some pushing and pressing tasks. The second way is to add an arm to the robot and use the gripper to grasp objects and perform more sophisticated loco-manipulation behaviors.

Within the first category, \cite{ji2022hierarchical,huang2023creating,jeon2023learning,arm2024pedipulate} propose a hierarchical framework to use the robot body and feet to generate interesting loco-manipulation behaviors. \cite{kim2022humanconquad} use teleoperation of human motion and employ DRL to re-target the motions to realizable quadrupedal loco-manipulation behaviors. \cite{cheng2023legs} use CL to learn different skills in simulation and then build a behavior tree to sequence the skills. 

Using only feet and body for manipulation is limited to simple loco-manipulation skills, as the robot is not able to grip/grasp objects/environments. To perform more sophisticated object manipulation, it is now common to mount a manipulator on quadrupeds. To perform loco-manipulation using quadruped with an arm, \cite{fu2023deep} use a privileged learning framework to train a policy that simultaneously deals with locomotion and manipulation. 
\cite{liu2024visual} develops a hierarchical framework, using a high-level policy that takes visual information as input and provides end-effector commands for a learned low-level controller. However, these approaches are limited to simple manipulation tasks with a quadruped and cannot reason about the complex loco-manipulation tasks. 
% Recently, 
% \cite{sleiman2023versatile} formulated the task and motion planning (TAMP) problem for quadrupedal loco-manipulation within the logic-geometric programming (LGP) framework \cite{toussaint2015logic}. They have shown interesting long-term reasoning loco-manipulation tasks using model-based control and planning techniques. However, there have not yet been many works from the learning community to perform a long sequence of loco-manipulation behaviors. 
Recently, \cite{kumar2023cascaded} proposed to use a library of loco-manipulation skill primitives and to compose them and generate a long sequence of loco-manipulation system. However, the manipulation part of the framework is still basic (such as opening a door). Using learning-based algorithms to perform a long sequence of complex loco-manipulation tasks is an interesting future frontier.

More recent work also explores how to learn loco-manipulation behaviors for bipedal robots, e.g.,
\cite{fu2024humanplus,dao2024sim,seo2023deep}. One advantage of using a humanoid form is the vast source of loco-manipulation data from human demonstration, either via motion capture, teleoperation, or video. However, bipedal loco-manipulation is also more challenging to achieve as compared to quadruped  due to quadruped loco-manipulation due to the latter being more stable. Loco-manipulation for humanoid robots is also an active area of research area and more research is to be done in the future.

\subsection{Foundation Models}\label{sec:foundation_models}
Learning locomotion skills and empowering robots with mobility is a key milestone towards building general-purpose robots that are truly useful. However, a notable gap still exists in terms of enabling robots to generalize their behaviors to arbitrary tasks, environments, and interactions with humans. The recent progress in training large-capacity pre-trained models with web-scale data (foundation models) has demonstrated impressive common-sense reasoning, learning, and perception capabilities without specialized training \cite{achiam2023gpt, touvron2023llama, team2023gemini}. Such capabilities are critical for creating general-purpose robots and have thus led to a plethora of works that combine robotics and foundation models in planning, reasoning, navigation, and simulation of robotic systems \cite{kira2022llmroboticspaperslist}. We refer readers to several existing survey papers summarizing these efforts \cite{hu2023toward,zhou2023language,firoozi2023foundation,xiao2023robot} and focus on discussing the ones applied to robot locomotion here. 

A common approach in combining foundation models with legged robots applies it for high-level planning \cite{lykov2024cognitivedog, xu2023creative}. In these methods, a foundation model is used to interpret the environment and task and to plan a sequence of skills that the robot should perform to achieve a longer horizon task. However, these methods usually do not have fine control of the robots' low-level behavior such as locomotion gaits. This is due to the fact that pre-trained foundation models are usually not trained with robotic data and cannot directly generate low-level robot actions. To leverage foundation models that also work with low-level locomotion control, researchers have designed different interfaces including foot contact patterns \cite{tang2023saytap}, and reward functions \cite{yu2023language, liang2024learning, ma2024dreureka} that allow foundation models to directly interact with robot's low-level controllers to perform tasks such as hopping, or high-five with a person. Alternatively, one may also fine-tune a large foundation model to directly output the low-level robot action using a large amount of robot data. This has been explored in the manipulation domain and shown promising generalization capabilities to novel tasks \cite{brohan2023rt}. Applying this idea to robot locomotion is a promising idea to obtain generalist robot locomotion controllers.

\section{Societal Impact}
Ever-more capable legged robots brings significant potential for both good and bad. This includes the potential to be deployed in many humanitarian applications, tackling dull-and-repetitive tasks in warehouse and agriculture operations, factory inspection, and performing jobs that are dangerous for humans, e.g., search and rescue and firefighting. At the same time, these robots have the potential to be used in multiple destructive applications. We believe that our community should play an active role in the discussions related to the regulations at this stage of development toward general-purpose robots (which are mostly legged systems) and remain alerted about the potential risks of this technology. 

Here, we aim to only list some immediate issues related to the use of general-purpose robots in the real world. 
While being preliminary, we believe initiatives from the robotics research community into these topics can have several positive impacts. In particular, we urge the audience of this paper to read the excellent article on the ethics of artificial intelligence and robotics \cite{muller2020ethics}. As well-said in the article: \emph{there is a tendency for businesses, the military, and some public administrations to “just talk” and do some “ethics washing” in order to preserve a good public image and continue as before. Actual policy is not just an implementation of ethical theory, but subject to societal power structures—and the agents that do have the power will push against anything that restricts them. There is thus a significant risk that regulation will remain
toothless in the face of economical and political power.} Within such a power structure, academia and researchers can constitute a powerful entity that raises awareness about the new technologies and their implications, and makes sure that the right policies are implemented.

One dangerous application is to use general-purpose robots as lethal autonomous weapon systems \cite{righetti2018lethal}. Unfortunately, we have seen such activities towards weaponizing quadrupeds in recent years.
Such systems would raise multiple important ethical and societal issues. From an ethical point of view, this question arises: should an autonomous system that \emph{does not suffer} or \emph{has no feelings} itself have the right to decide to kill a person? on the legal side, who would be responsible (and punished) for the potential mistakes of an autonomous weapon system? Also, from the human rights perspective, autonomous weapons would not be aligned with international humanitarian law. Human Rights Watch and Harvard Law School’s International Human Rights Clinic (IHRC) notes \cite{docherty2012losing}: 
\emph{... such revolutionary weapons would not be consistent with international humanitarian law and would increase the risk of death or injury to civilians during armed conflict. A preemptive prohibition on their development and use is needed.}.
A recent open letter \cite{letter2022generalpurpose} signed by multiple companies (Agility Robotics, ANYbotics, Clearpath Robotics, Open Robotics, Unitree, Boston Dynamics) aligns with this view and condemns any use of weaponized mobile robots. Activities of this kind, together with practical measures, can help to safe guard the technology of legged robots in the future.

Another impact of developing general-purpose robots in the near future is the possible human jobs lost due to new automation abilities \cite{pham2018impact}. Replacing humans in health-threatening jobs, e.g., firefighting, search and rescue in the wild, dangerous industrial work, and care-giving in pandemics \cite{shen2020robots}, can be of significant benefit. However, the problem arises with the potential to replace human workers by robots for many other job categories, and the impact on the lives of displaced workers. An alternative that is regularly advocated is to use robots to take over the physical work while humans use their cognitive capabilities. Human-robot collaboration is another scenario that is advocated over replacing humans. While human history has seen significant shifts due to mechanization before, as with agriculture, for example, past outcomes and adaptations may be a poor predictor of future outcomes on these issues. A recent study spanning different cultures and industries suggests that increased exposure to robots can lead to increased job insecurity \cite{yam2023rise}. With rapidly advancing technology, it will be important for policy makers to consider the impact of robotics on the society and economy of the future. 

Environmental concerns currently see minimal discussion in the robotics community. As deep learning is increasingly used with models of increasing size, it is important to discuss the carbon footprint of training and using these massive models. This issue is already discussed in the machine learning community, and it is important that the robotics community also thinks about how to mitigate the rapid and unsustainable growth of compute requirements. Another more nuanced environmental issue is that of the life-cycle of the robot itself: what happens at the end of a robot’s operational life? Currently legged robots and humanoids are not contributing significantly in this area and the question is more immediate concern for industrial robots and autonomous cars. However, the adoption of quadrupeds and humanoids on a larger scale means that we may eventually see large numbers of these robots that require disposal or recycling when they are no longer functional. It is therefore incumbent on us to think about this issue sooner rather than later.

We have only touched on only some of the immediate risks related to general-purpose robotic systems. This list is by no means complete.  It does not take other aspects of the problem into account that are related to the software of an embodied AI, including enabling unwanted surveillance, biases in decision making, possible amplification of existing inequities, and more.

\section{Conclusions}
Learning-based methods are at the heart of so many of the advances seen in legged locomotion control. In this article, we summarized the core issues and methods for quadrupedal locomotion learning, as well as touching on recent related methods for bipedal locomotion. 
While it is impossible to be fully comprehensive, we hope that this survey provides a valuable framework for understanding recent progress along with ample points of reference.
We have further outlined unsolved problems and future directions that, because of their difficulty and broad nature, will provide significant challenges for years to come. 
However, it is reasonable to expect the rapid pace of innovation to continue, yielding ever-more capable legged locomotion for quadrupeds and bipeds.

%% Michiel:  I reworded much of the conclusions

%% Michiel:  commented out the acknowledgements for now
% \begin{acks}
% This class file.
% \end{acks}

\bibliographystyle{abbrvnat}
\bibliography{references}

\end{document}

%% file: defs.tex
\newcommand{\E}[2]{\operatorname{\mathbb{E}}_{#1}\left[#2\right]}
\newcommand{\sspace}{\mathcal{S}}
\newcommand{\ospace}{\mathcal{O}}
\newcommand{\aspace}{\mathcal{A}}

\newcommand{\state}{\mathbf{s}}

\newcommand{\st}{{\state_t}}

\newcommand{\stp}{{\state_{t+1}}}

\newcommand{\action}{\mathbf{a}}

\newcommand{\at}{{\action_t}}